%% file: main.tex
\pdfoutput=1

\documentclass[11pt]{article}

\usepackage[]{ACL2023}
\usepackage{xcolor}
\definecolor{darkgreen}{rgb}{0.0, 0.5, 0.0}
\definecolor{lightblue}{RGB}{173,216,230}
\definecolor{lightred}{RGB}{255,182,193}
\definecolor{lightgreen}{RGB}{173,255,47}
\definecolor{lightyellow}{RGB}{255,255,204}
\definecolor{violet}{RGB}{90, 19, 242}

\usepackage{times}
\usepackage{latexsym}
\usepackage{graphicx}
\usepackage{subcaption}
\usepackage{hyperref}
\usepackage{wrapfig}
\usepackage{url}
\usepackage{graphicx}
\usepackage{multirow}
\usepackage{hyperref}
\usepackage{booktabs}
\usepackage{longtable}
\usepackage{longtable}
\usepackage{tabularx}
\usepackage{booktabs}
\usepackage{siunitx}
\usepackage{hyperref}
\usepackage{enumitem}
\usepackage{amsmath}
\usepackage{xcolor}
\usepackage{tcolorbox}
\usepackage{soul}
\usepackage{makecell}
\usepackage{multirow}
\usepackage{booktabs}
\usepackage{graphicx}
\usepackage{amsmath}
\usepackage{amssymb}
\usepackage{placeins}
\usepackage[titletoc]{appendix}
\usepackage{xltabular}
\usepackage{amsmath}
\DeclareMathOperator*{\argmin}{arg\,min}
\usepackage[T1]{fontenc}

\usepackage[utf8]{inputenc}

\usepackage{microtype}

\usepackage{inconsolata}

\title{Towards Knowledge Checking in Retrieval-augmented Generation: 

A Representation Perspective  }

\author{Shenglai Zeng$^{1}$\thanks{Work done during his internship at Amazon Search. }, Jiankun Zhang, Bingheng Li$^1$, Yuping Lin$^1$, Tianqi Zheng$^{2}$, Dante Everaert$^{2}$ \\  \textbf{Hanqing Lu$^{2}$, Hui Liu$^2$, Hui Liu$^1$, Yue Xing$^1$, Monica Xiao Cheng$^2$, Jiliang Tang$^1$ } \\ 
$^1$Michigan State University  \quad $^2$ Amazon.com   
  \\
\{zengshe1, libinghe, linyupin, liuhui7, xingyue1,  tangjili\}@msu.edu, \\
\{tqzheng, danteev, luhanqin,liunhu, chengxc\}@amazon.com
}

\begin{document}
\maketitle
\newtheorem{definition}{Definition}
\input{Sections/Abstract}

\input{Sections/Introduction}

\input{Sections/Related_Works}

\input{Sections/Methods}

\input{Sections/Experiment}

\input{Sections/Conclusions}
\input{Sections/Limitations}
\bibliography{anthology}

\input{Sections/Appendix}
\label{sec:appendix}

\end{document}

%% file: Sections/Abstract.tex
\begin{abstract}
\label{abstract}
Retrieval-Augmented Generation (RAG) systems have shown promise in enhancing the performance of Large Language Models (LLMs). However, these systems face challenges in effectively integrating external knowledge with the LLM's internal knowledge, often leading to issues with misleading or unhelpful information. This work aims to provide a systematic study on knowledge checking in RAG systems. We conduct a comprehensive analysis of LLM representation behaviors and demonstrate the significance of using representations in knowledge checking. Motivated by the findings, we further develop representation-based classifiers for knowledge filtering. We show substantial improvements in RAG performance, even when dealing with noisy knowledge databases. Our study provides new insights into leveraging LLM representations for enhancing the reliability and effectiveness of RAG systems.
\end{abstract}

%% file: Sections/Introduction.tex



\section{Introduction}
\label{Intro}

Retrieval-augmented generation (RAG) is a technique designed to enhance the outputs of large language models (LLMs) by incorporating relevant information retrieved from external knowledge sources. This approach has been applied to various domains and scenarios \cite{liu2022llama,chase2022langchain,van2023clinical,ram2023context,shi2023replug,siriwardhana2023improving,parvez2021retrieval,panagoulias2024augmenting,pipitone2024legalbench,mozharovskii2024evaluating}. 
It typically operates in two stages: retrieval and generation. In the retrieval stage, relevant knowledge from an external database is retrieved based on the user query. Then, in the generation stage, the retrieved information is integrated with the query to form an input for LLMs to generate responses. 

In RAG, two potential knowledge sources can be utilized to answer input queries: LLM's internal knowledge and the external knowledge provided in the context. Ideally, these external and internal knowledge sources should be effectively integrated. However, existing works have shown that LLMs often struggle to identify the boundaries of their own knowledge and tend to prioritize external information over their internal knowledge learned during pre-training~\cite{ren2023investigating,tan2024blinded,wang2023resolving,ni-etal-2024-llms,liu2024ra,wang2023self,zeng2024good}. This characteristic can potentially degrade the generation quality of RAG when the quality of external knowledge is low. On one hand, the external knowledge may be \textbf{misleading} \cite{zou2024poisonedrag,dengpandora}. For instance, \citet{zou2024poisonedrag} proposed the PoisonedRAG approach, demonstrating that LLMs can be easily manipulated into producing incorrect information simply by injecting false answers corresponding to targeted queries into the retrieval database. On the other hand, although some retrieved contexts are semantically similar to a query, they may only \textit{superficially
related to the topic but lack the answer} to the question\cite{yoranmaking,fang-etal-2024-enhancing}. Such contexts can distract LLMs and consequently hurt RAG performance.

Thus, it is important to conduct knowledge checking in RAG systems. To achieve this goal,  we design the following critical tasks:
\begin{enumerate}[label=(\alph*), itemsep=0pt, parsep=0pt]
    \item \textbf{Internal Knowledge Checking:} When a user inputs a query, the LLM should first check whether it possesses internal knowledge relevant to the query, i.e., { Internal Knowledge Checking ({\bf Task 1})}. This task serves as a foundation for subsequent checks.
    \item \textbf{Helpfulness Checking}: Helpfulness checking is to examine if the external knowledge is helpful\footnote{"Helpfulness" here refers to the relevance of information to the query, information directly addressing the question is considered helpful.} to answer the input query. We design {Informed Helpfulness Checking ({\bf Task 2})} when the LLM has internal knowledge about the query and {Uninformed Helpfulness Checking ({\bf Task 3})} when the LLM lacks internal knowledge about the query. As as an extreme case of Task 2, we design { Contradiction Checking ({\bf Task 4} )} to check if internal knowledge has any contradictions with the retrieved external information. 
    
\end{enumerate}

\noindent A straightforward approach to tackle these tasks can directly prompt LLMs\cite{asaiself,wang2023self,liu2024ra,zhang2024retrievalqa}. Alternatively, we could examine superficial indicators of LLMs, such as probability scores~\cite{wang2024self,jiang2023active} or perplexity~\cite{zou2024poisonedrag}. However, based on our evaluation in Section \ref{Sec: Representation}, we find that none of these methods can effectively accomplish these tasks. 



Recent studies~\cite{zou2023representation,lin2024towards,zheng2024prompt} have shown that LLMs' representations exhibit distinct patterns when encountering contrasting high-level concepts, such as harmful versus harmless prompts  . This observation prompts us to investigate \textit{whether LLMs' representations also display distinct behaviors and can be leveraged in knowledge checking tasks?} To answer this question, we conduct a comprehensive study and analysis of LLM representation behaviors regarding the aforementioned tasks, including PCA-based checking as well as contrastive-learning-based checking (Section \ref{rep_method}). Our analysis reveals that positive and negative samples exhibit  different behaviors in the representation space. Consequently, representation-based methods demonstrate significantly superior performance in the aforementioned tasks. Leveraging these findings, we utilize representation classifiers for knowledge filtering. Results show that simple filtering of contradictory and irrelevant information substantially improves RAG performance, even in scenarios with poisoned knowledge databases.

%% file: Sections/Related_Works.tex
\section{Related Work}
\label{Intro}

\subsection{Robustness Issues in RAG}


RAG faces robustness challenges. A growing body of research~\cite{ren2023investigating,tan2024blinded,wang2023resolving,ni-etal-2024-llms,liu2024ra,wang2023self,zeng2024good} has revealed that LLMs often struggle to identify their knowledge boundaries, tending to over-rely on provided context. This vulnerability makes RAG susceptible to failure with misleading~\cite{zou2024poisonedrag,dengpandora,xieadaptive} or unhelpful context~\cite{yoranmaking,asaiself,liu2024ra}. 
\subsection{Knowledge Checking in RAG}

Recent research has explored various knowledge checking tasks in RAG systems to address the aforementioned issues. Some studies leverage LLMs' self-generated responses to determine whether a question is answerable without external information~(\textbf{answer-based methods}). \cite{ren2023investigating,liu2024ra,asaiself,zhang2024retrievalqa,wang2024self,jeong-etal-2024-adaptive} or to assess the relevance of retrieved context \cite{liu2024ra,asaiself}. Other approaches employ explicit metrics such as probability \cite{wang2024self,jiang2023active} to evaluate the necessity of retrieval, or perplexity \cite{zou2024poisonedrag} to judge the reliability of context~(\textbf{probability-based methods}). 


\subsection{Representation Engineering on LLMs}

Recent studies have shown that LLMs' representation space contains rich information for analyzing and controlling their high-level behaviors. \citet{zou2023representation} introduced RepE techniques, demonstrating that projecting representations onto a 'reading vector' can reveal safety-related aspects, aspects such as honesty, confidence~\cite{liu2024ctrla} and harmlessness. Subsequent research by \citet{zheng2024prompt} and \citet{lin2024towards} also indicates harmful and harfulness prompts are naturally distinguishable in the representation space. 



%% file: Sections/Methods.tex
\section{Representations for Knowledge Checking}
\label{Sec: Representation}
Drawing on insights from cognitive neuroscience, previous studies~\cite{zou2023representation,zheng2024prompt,lin2024towards} have demonstrated the potential of using LLMs' representation to indicate contrast high-level concepts. In this subsection, we investigate whether LLMs' representations also show distinct patterns in knowledge checking tasks and can therefore be used to improve their performance. We begin by introducing our representation-based checking procedures in Section \ref{rep_method}, which includes both PCA-based checking(\textit{rep-PCA}) and the contrastive-learning-based checking(\textit{rep-con}).  We then visualize and compare the performance of our representation-based methods against traditional approaches across four knowledge checking tasks from Section~\ref{Sec:query} to Section~\ref{Sec:conflict}.


\label{methods}

\subsection{Representation-based Knowledge Checking}
\label{rep_method}
\begin{table*}[!htpb]
\centering
\caption{Performance comparison of different methods on RAG robustness aspects}
\label{tab:rag_robustness}
\resizebox{\textwidth}{!}{
\begin{tabular}{@{}l|cccc|cccc|cccc|cccc@{}}
\toprule
 & \multicolumn{4}{c|}{Internal Knowledge} & \multicolumn{4}{c|}{Uninformed Helpfulness} & \multicolumn{4}{c|}{Informed Helpfulness} & \multicolumn{4}{c}{Conflict Detection} \\
\cmidrule(l){2-5} \cmidrule(l){6-9} \cmidrule(l){10-13} \cmidrule(l){14-17}
Method & Acc & Pre & Rec & F1 & Acc & Pre & Rec & F1 & Acc & Pre & Rec & F1 & Acc & Pre & Rec & F1 \\
\midrule
DIRECT & 0.47 & 0.51 & 0.76 & 0.61 & 0.55 & 0.53 & 0.97 & 0.69 & 0.56 & 0.53 & 0.99 & 0.69 & 0.50 & 0.50 & 0.99 & 0.66 \\
ICL & 0.54 & 0.56 & 0.77 & 0.65 & 0.55 & 0.53 & 0.98 & 0.69 & 0.55 & 0.53 & 1 & 0.69 & 0.42 & 0.45 & 0.79 & 0.58 \\
COT & 0.49 & 0.53 & 0.78 & 0.63 & 0.68 & 0.62 & 0.94 & 0.75 & 0.68 & 0.61 & 0.97 & 0.75 & 0.41 & 0.45 & 0.81 & 0.58 \\
Self-RAG(Mistral) & 0.47 & 0.51 & 0.69 & 0.59 & 0.63 & 0.57 & 0.96 & 0.72 & 0.60 & 0.55 & 0.98 & 0.71 & - & - & - & -\\
Prob(Lowest)& 0.69 &0.69  & 0.77 & 0.73 &0.62  & 0.60 & 0.74 & 0.66 & 0.60 & 0.57 & 0.79 & 0.66 &0.50  &  0.50& 1.00 &0.67  \\
Prob(Avg) & 0.65 & 0.68 & 0.69 &0.69  &0.61  & 0.60 &0.65  & 0.62 & 0.60 & 0.58 & 0.68 &0.63  & 0.50 &0.50  & 1.00 &0.67  \\
Perplexity & 0.55 & 0.55 & 0.98 & 0.71 & 0.50 & 0.50 & 1.0&0.67  & 0.50 & 0.50 &1.00  & 0.67 & 0.50 & 0.50 & 1.0 & 0.67 \\
\textit{Rep-PCA(Mistral)} & 0.75 & 0.72 & 0.81 & 0.76 & 0.79 & 0.77 & 0.81 & 0.79 & 0.81 & 0.80 & 0.81 & 0.81 & 0.91 & 0.92 & 0.90 & 0.91 \\
\textit{Rep-Con(Mistral)} & 0.78 & 0.72 & 0.86 & 0.78 & 0.81 & 0.80 & 0.82 & 0.81 & 0.85 & 0.84 & 0.85 & 0.85 & 0.95 & 0.91 & 0.99 & 0.95 \\
\bottomrule
\end{tabular}
}
\vspace{-0.2in}
\end{table*}

\paragraph{Problem formulation.}
In this subsection, we aim to analyze and classify the internal representation behavioral differences of LLMs for above-mentioned knowledge checking tasks when confronted with various types of inputs. To achieve this, we propose training a classifier to distinguish LLMs' internal behaviors based on their representations. Our main analysis uses Mistral-7B-Instruct-v0.1\cite{jiang2023mistral} as the LLM, focusing on the last input token's representations in the final layer. \footnote{Ablation studies on \textbf{other layers and models} are presented in Appendix \ref{app:other_layer} and \ref{app:other_model}, respectively.} Following \cite{zou2023representation}, we use both positive(e.g. queries with knowledge) and negative (e.g. queries without knowledge) samples as inputs, collecting the corresponding internal representations. Specifically, let $V^+ = \{v_i^+, c^+\}_{i=1}^{N^+}, \: V^- = \{v_j^-,c^-\}_{j=1}^{N^-}$represent the internal representations of positive and negative samples and corresponding labels, respectively.  The classifier is trained to differentiate between these samples, corresponding to various LLM behaviors. The construction of positive/negative samples in different tasks is shown in Appendix \ref{App:rep-prompts} and Table \ref{tab:rep-scenarios-checking}. Next, we introduce two methods to implement knowledge checking.

\begin{figure*}[t]
\centering
\resizebox{\textwidth}{!}{%
    \begin{minipage}{\textwidth}
        \subfloat[Internal Knowledge ]{\includegraphics[width=.25\textwidth]{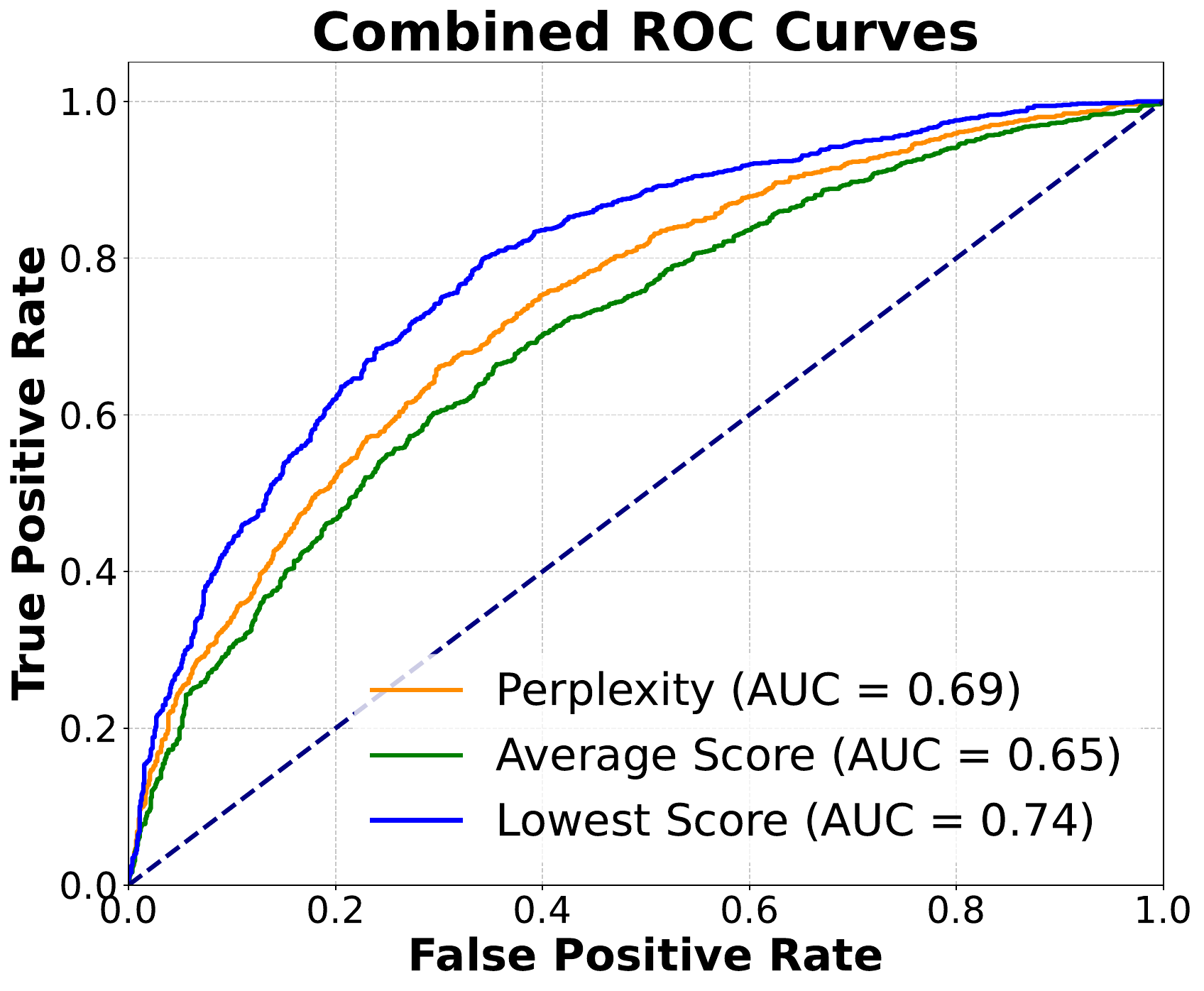}
        \label{fig:roc_query}}
        \subfloat[Uninformed helpfulness]{\includegraphics[width=.25\textwidth]{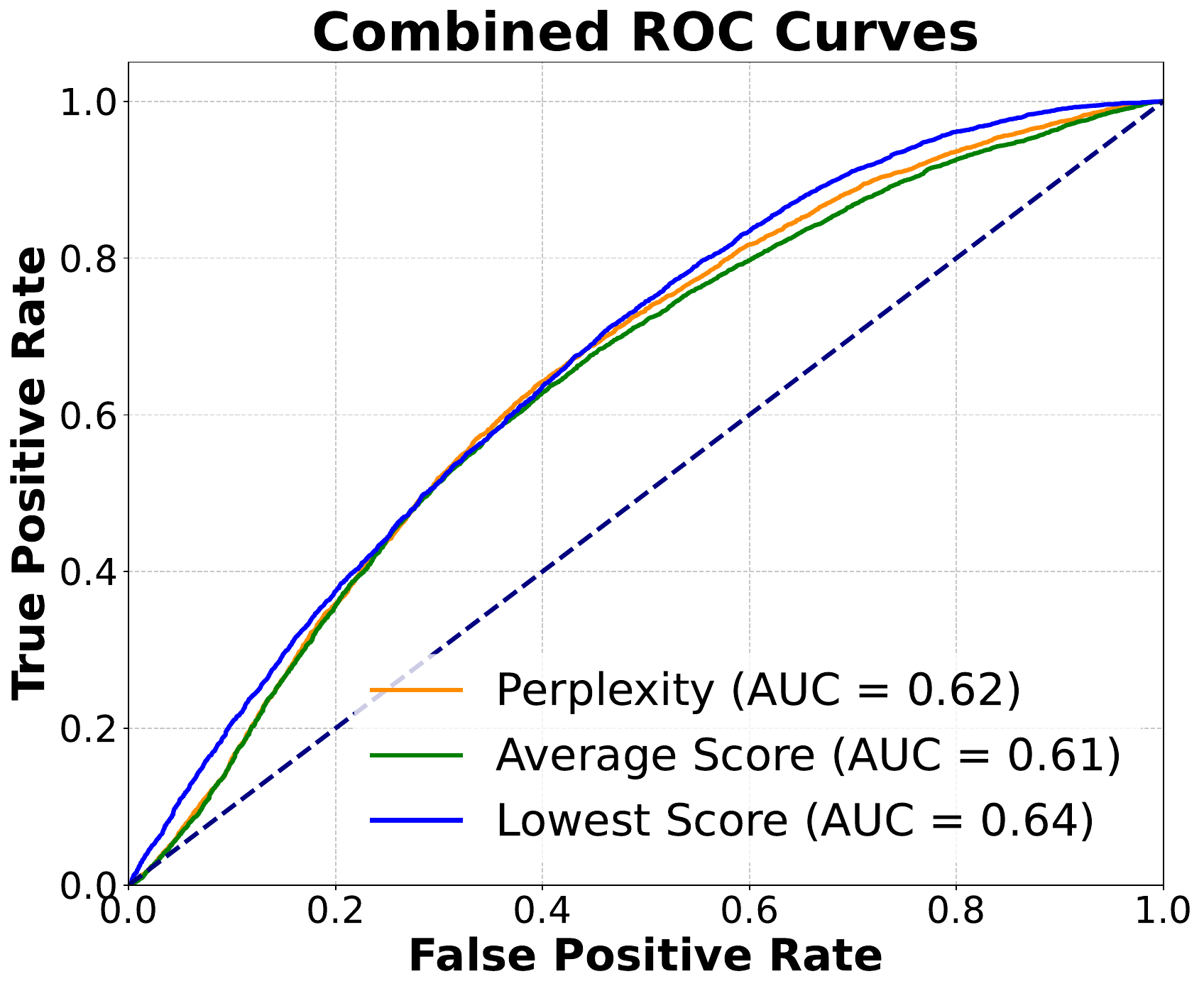}
        \label{fig:roc_uninfor_help}}
        \subfloat[Informed helpfulness]{\includegraphics[width=.25\textwidth]{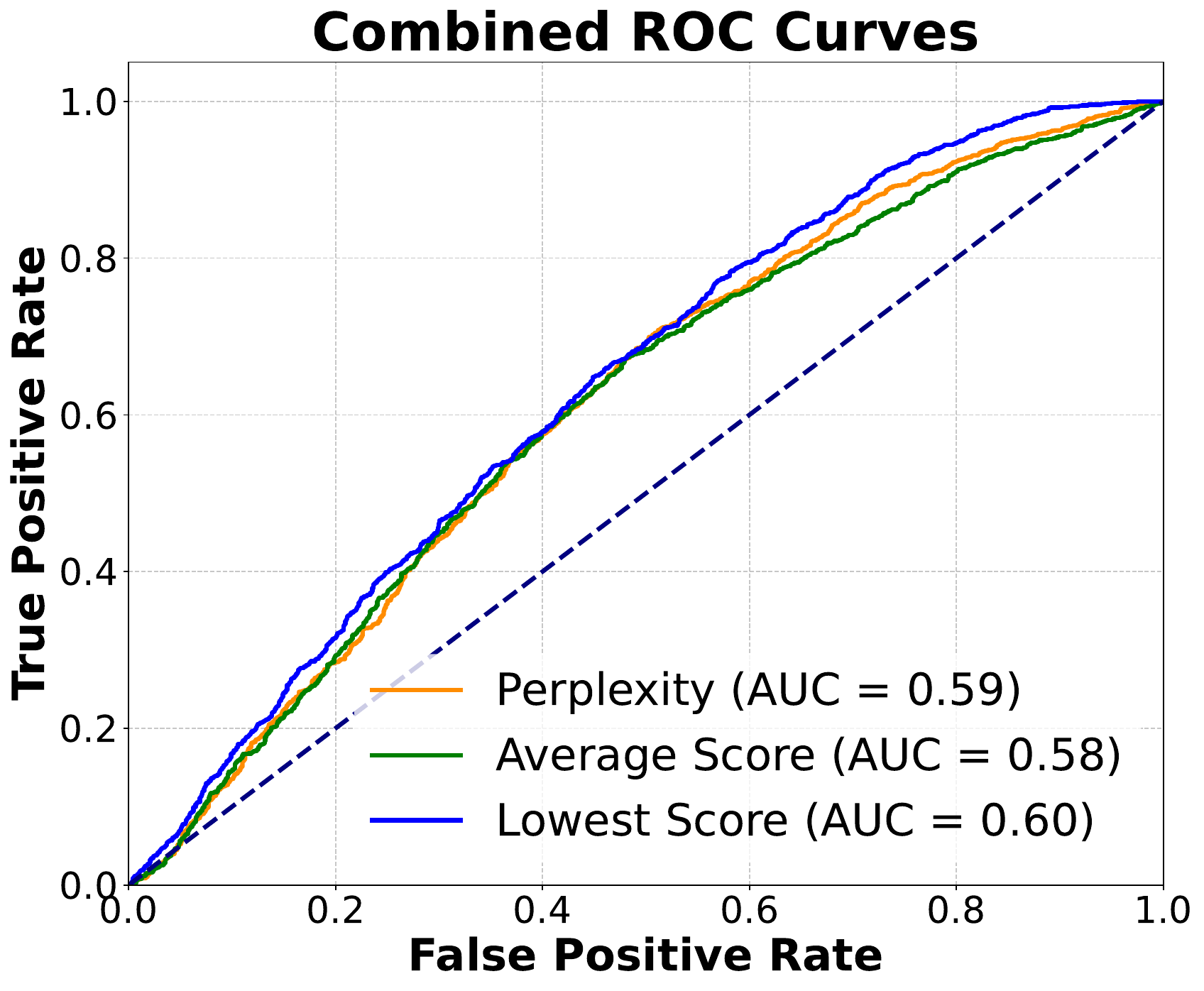}
        \label{fig:roc_infor_help}}
        \subfloat[Contradiction]{\includegraphics[width=.25\textwidth]{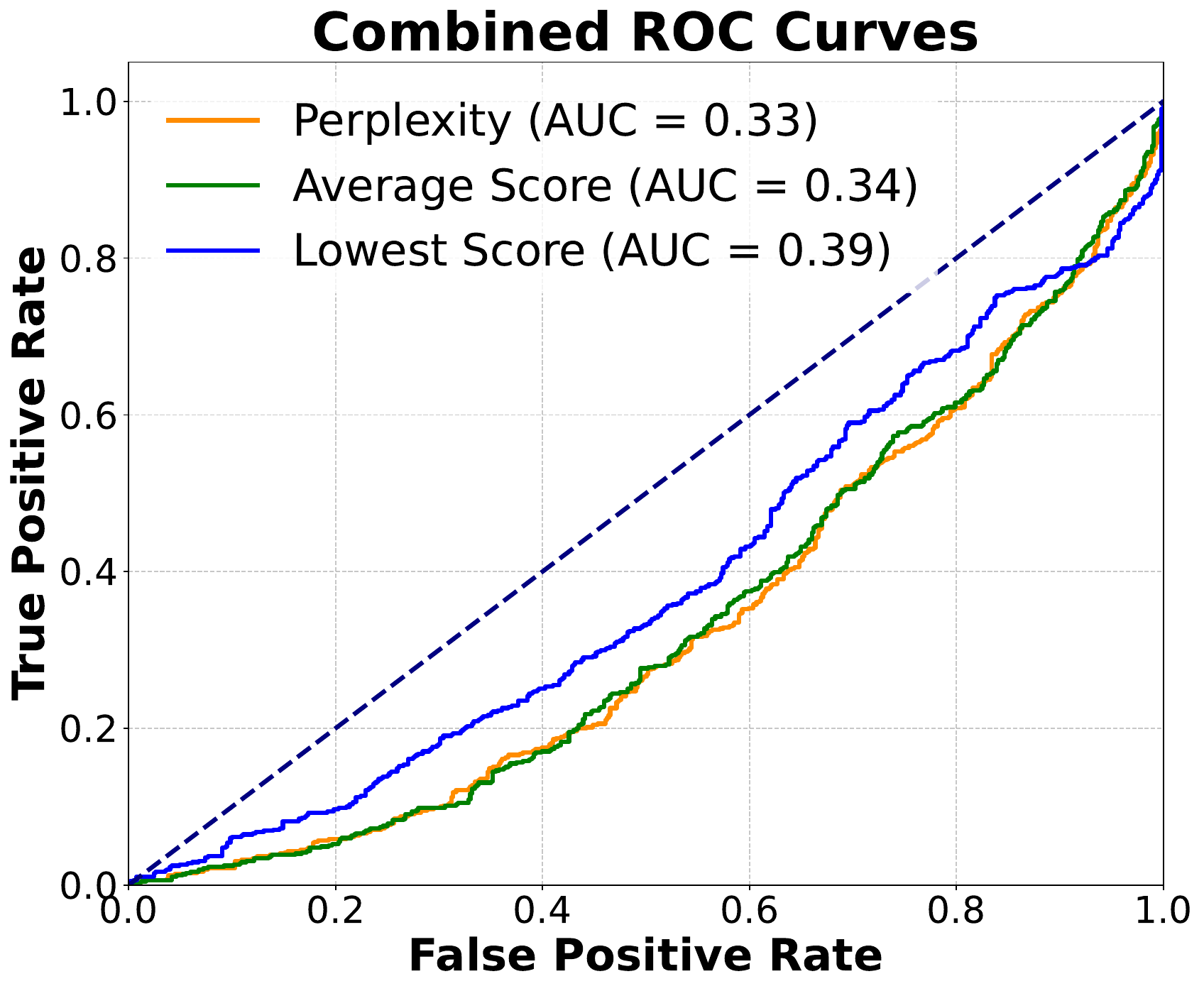}
        \label{fig:roc_contra}}
        
    \end{minipage}
}

\caption{ROC curve of probability-based methods}

\label{fig:roc}
\end{figure*}

\begin{figure*}[t]
\centering
\resizebox{\textwidth}{!}{%
    \begin{minipage}{\textwidth}
        \subfloat[Internal Knowledge ]{\includegraphics[width=.25\textwidth]{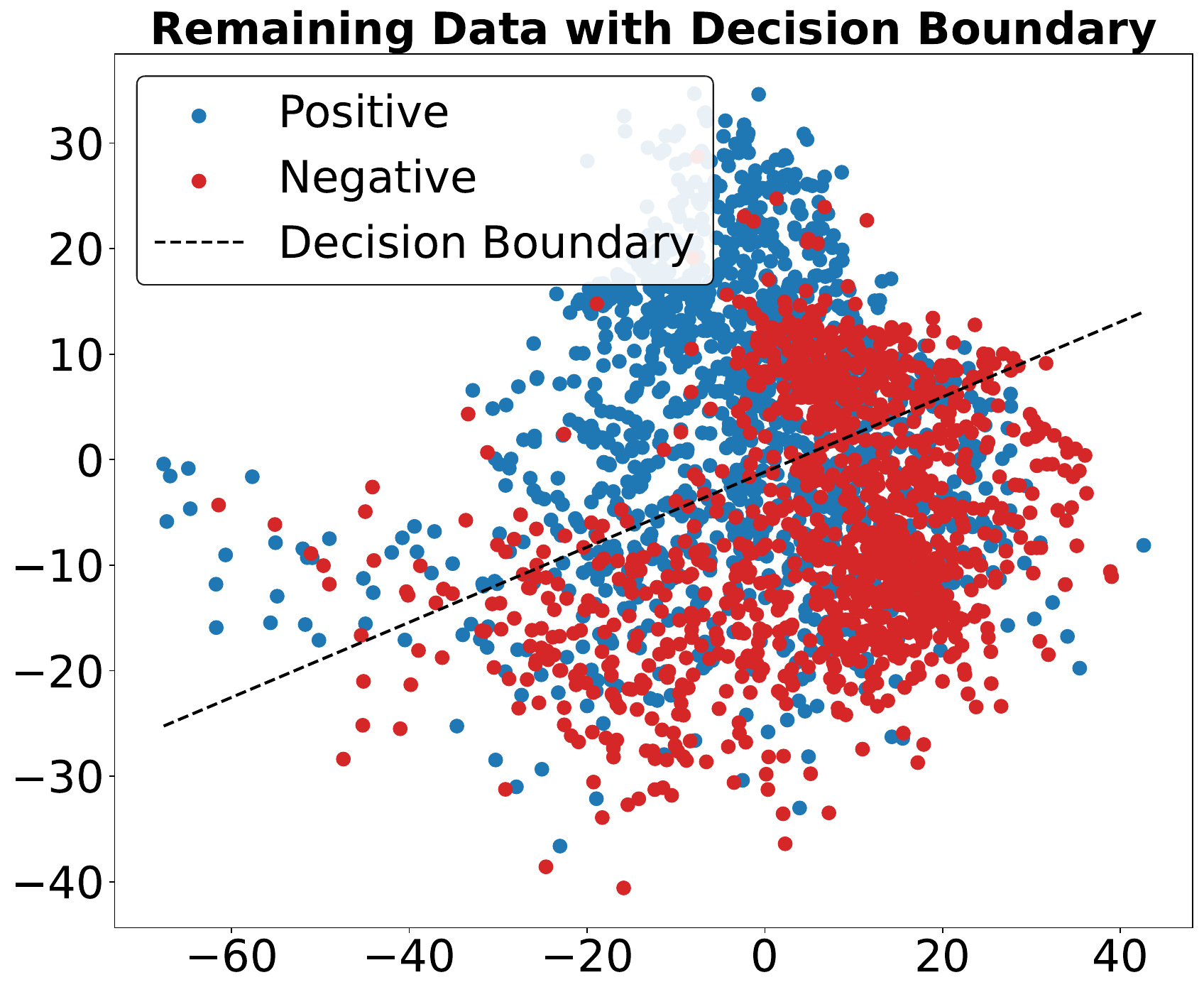}
        \label{fig:PCA_query}}
        \subfloat[Uninformed helpfulness]{\includegraphics[width=.25\textwidth]{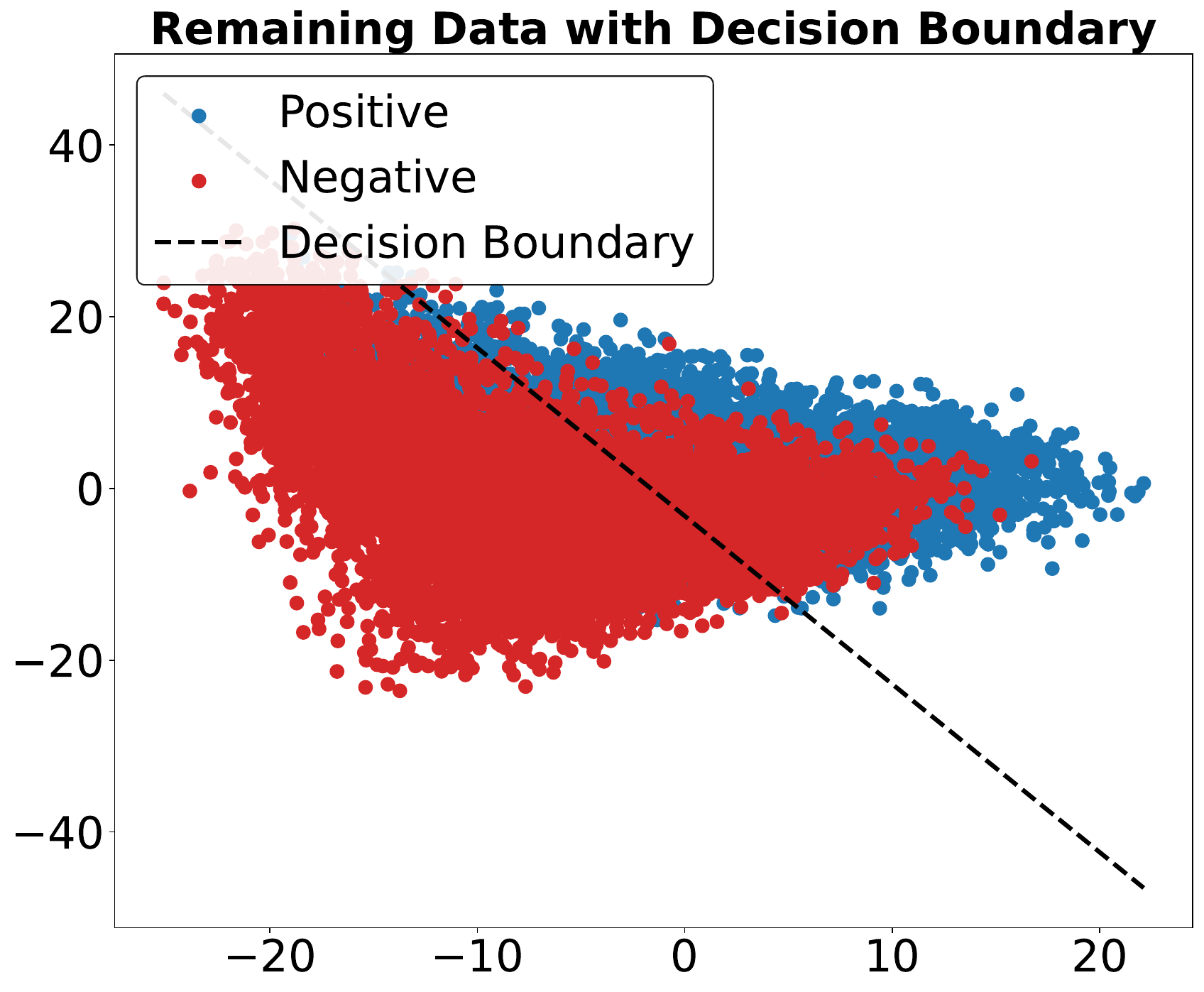}
        \label{fig:PCA_unknown_helpful}}
        \subfloat[Informed helpfulness]{\includegraphics[width=.25\textwidth]{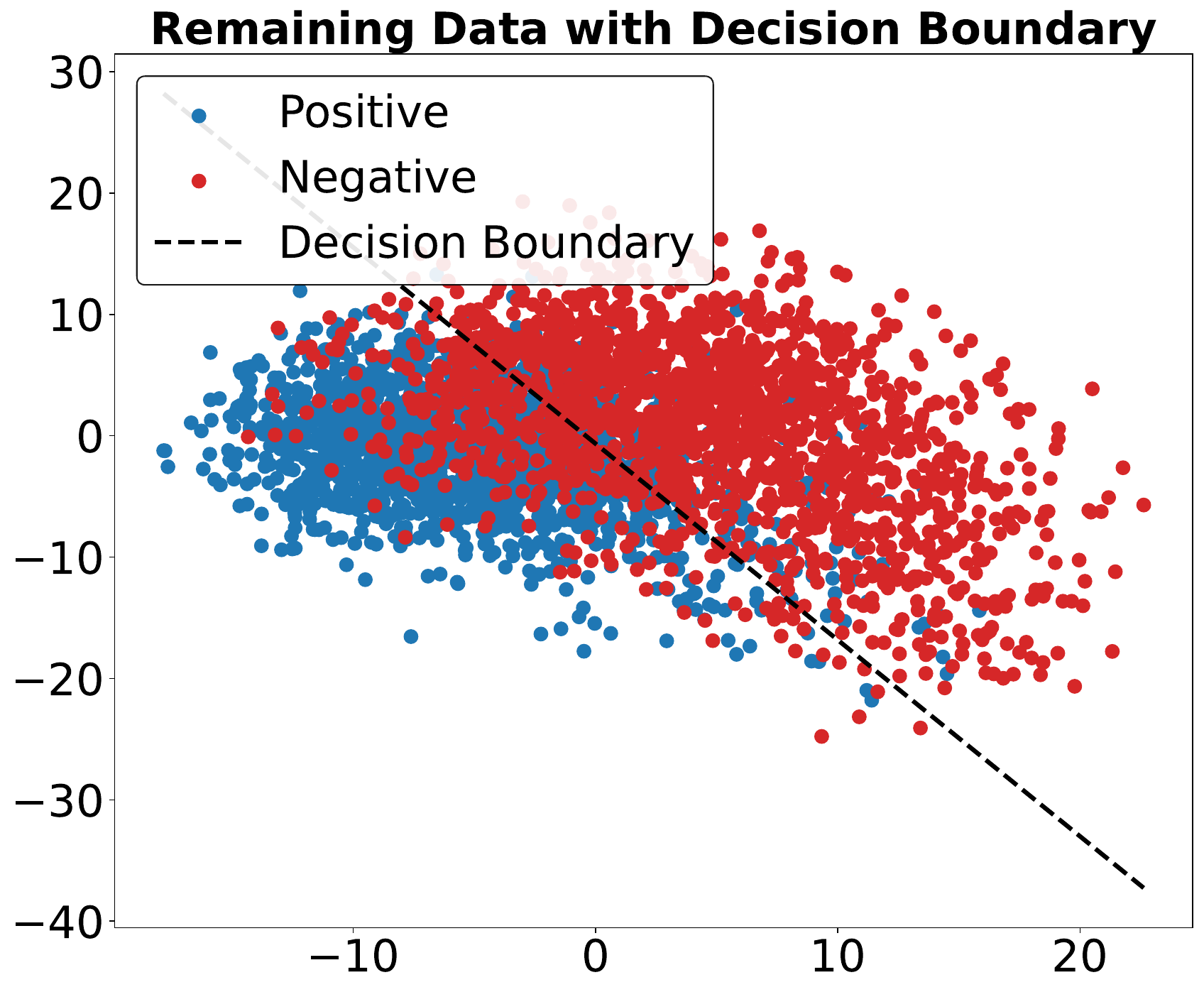}
        \label{fig:PCA_known_helpful}}
        \subfloat[Contradiction]{\includegraphics[width=.25\textwidth]{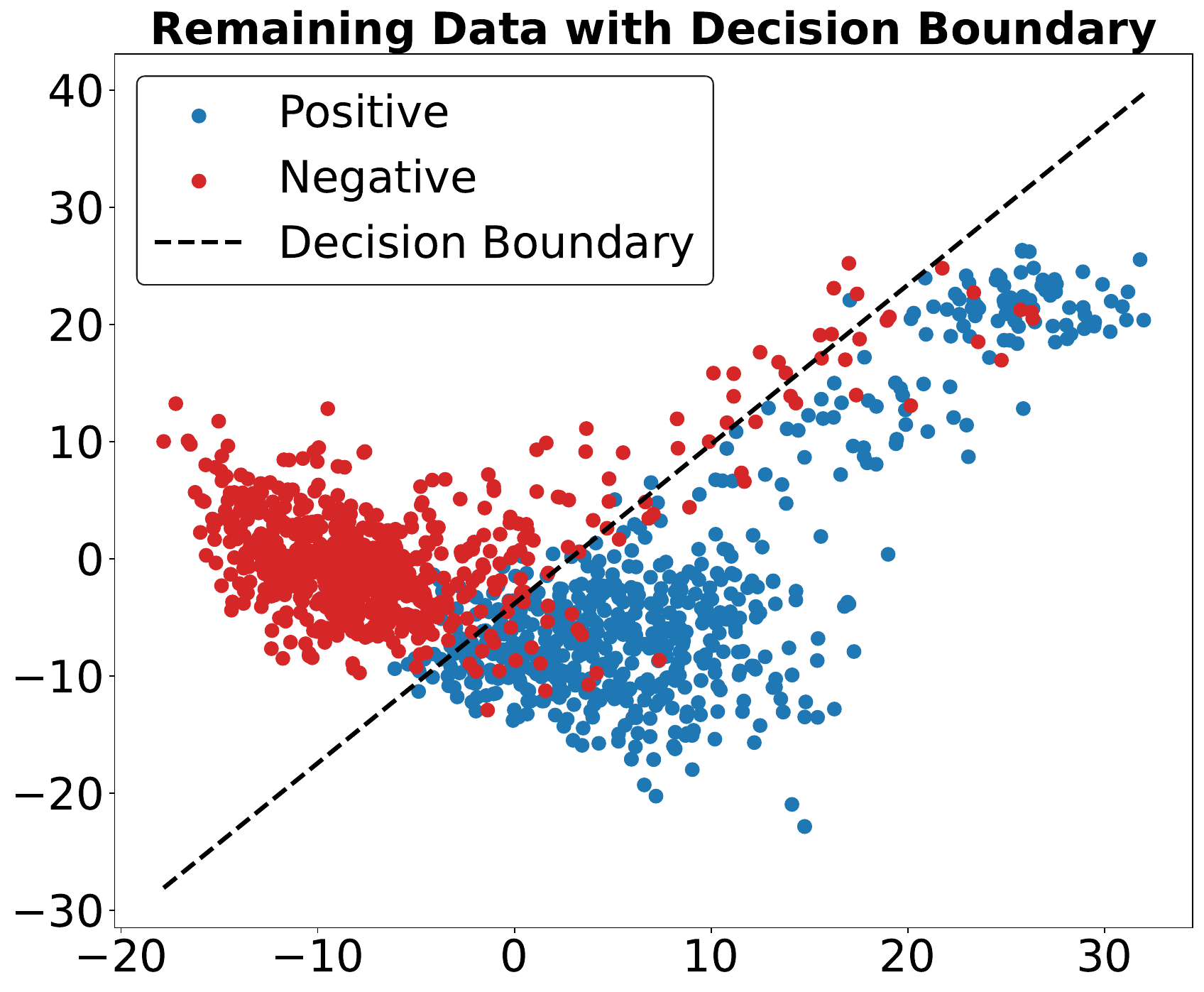}
        \label{fig:PCA_contra}}
        
    \end{minipage}
}

\caption{Visualization on PCA space}
\label{Model Choice}

\end{figure*}

\begin{figure*}[htbp]
\centering
\resizebox{\textwidth}{!}{%
    \begin{minipage}{\textwidth}
        \subfloat[Internal Knowledge ]{\includegraphics[width=.25\textwidth]{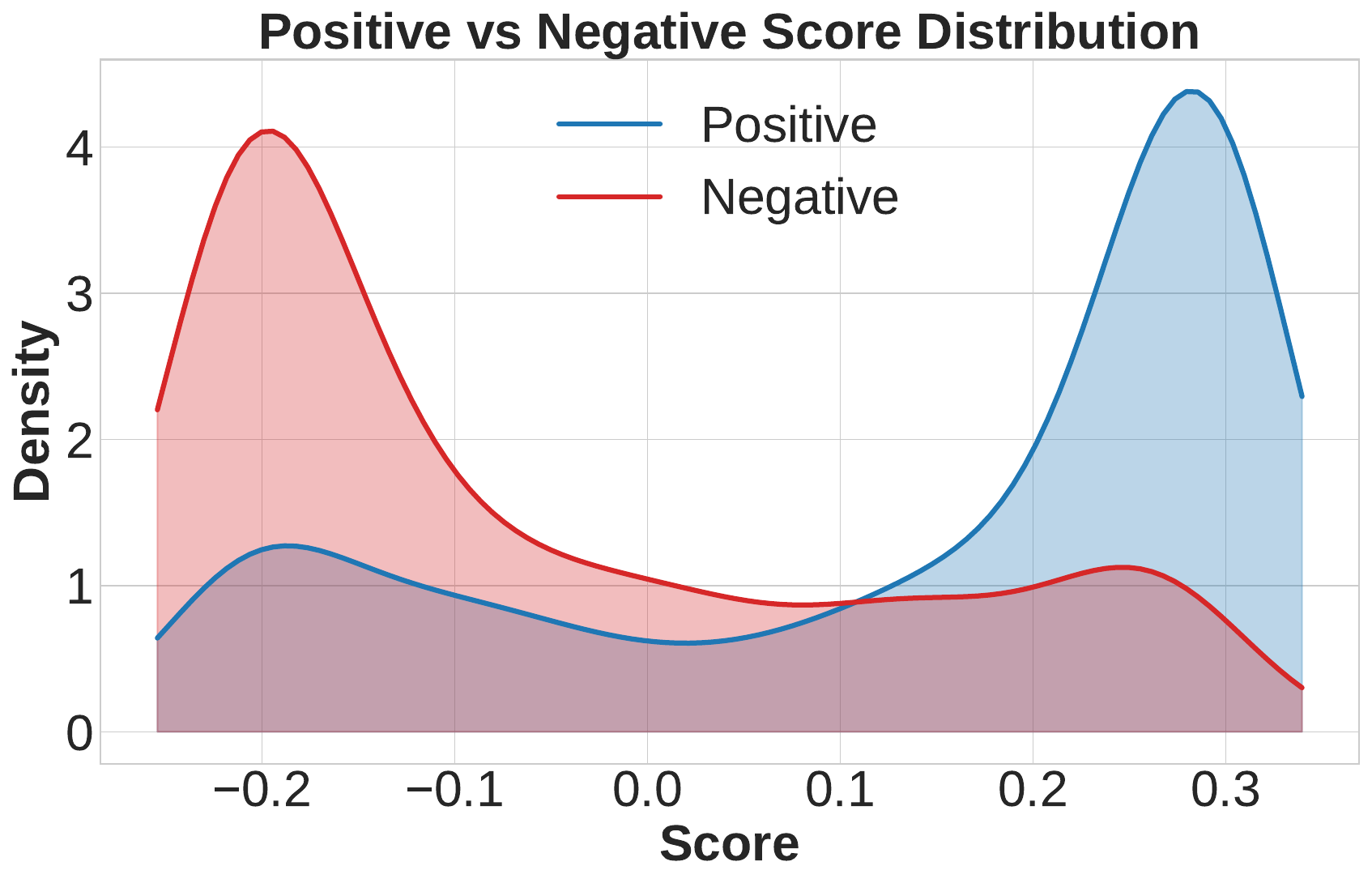}
        \label{fig:con_query}}
        \subfloat[Uninformed helpfulness]{\includegraphics[width=.25\textwidth]{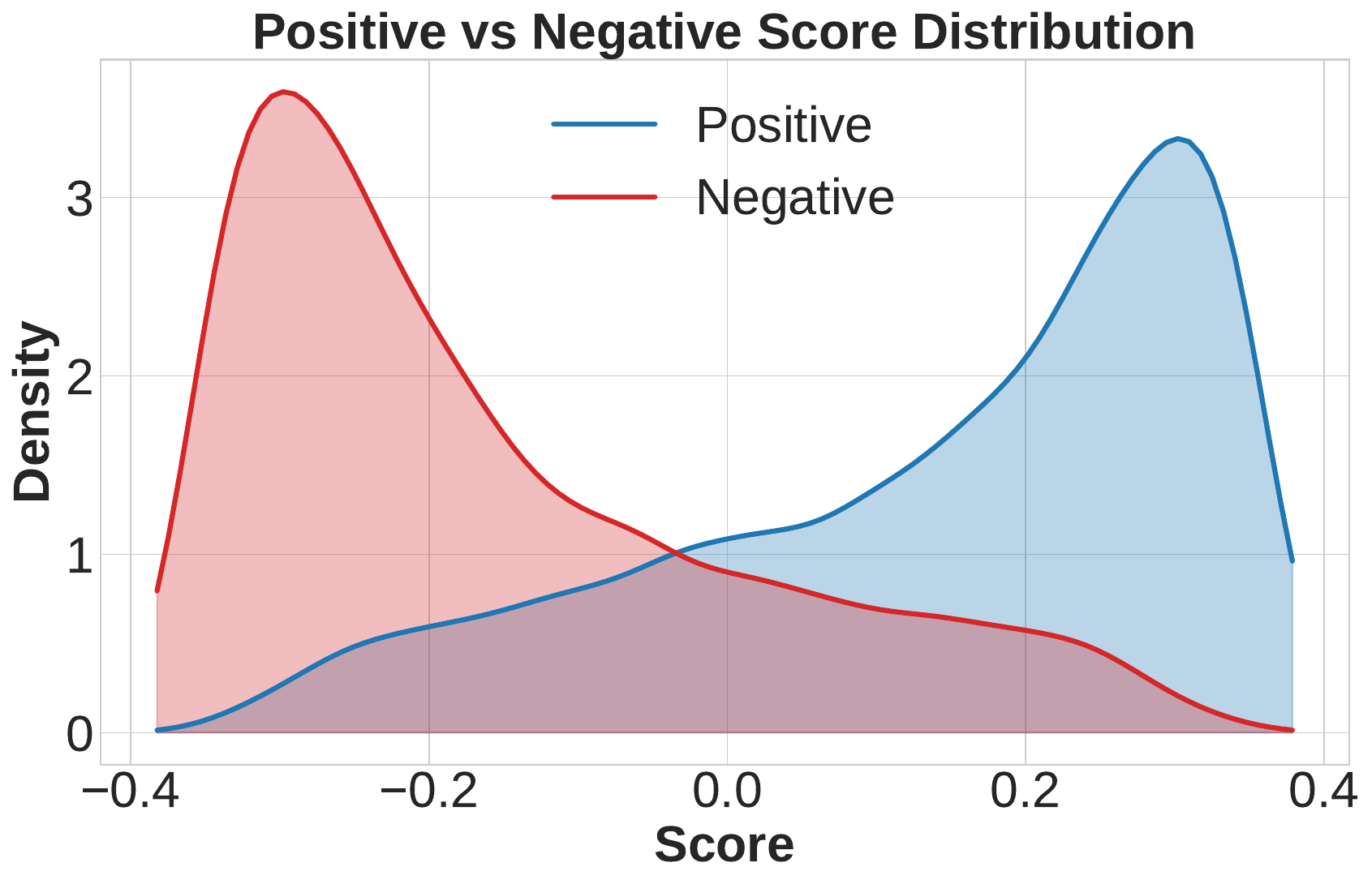}
        \label{fig:con_unknown_help}}
        \subfloat[Informed helpfulness]{\includegraphics[width=.25\textwidth]{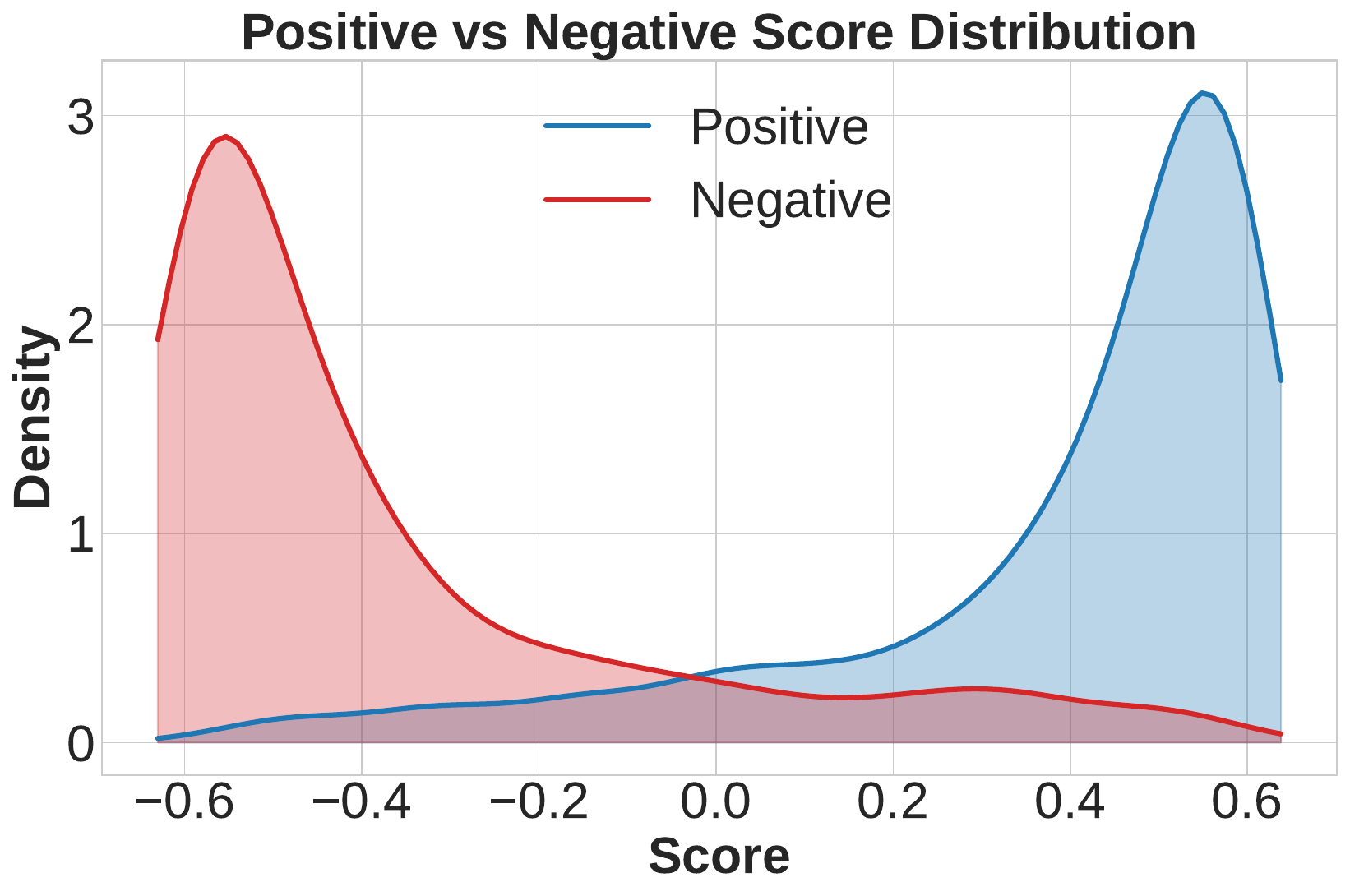}
        \label{fig:con_known_help}}
        \subfloat[Contradiction]{\includegraphics[width=.25\textwidth]{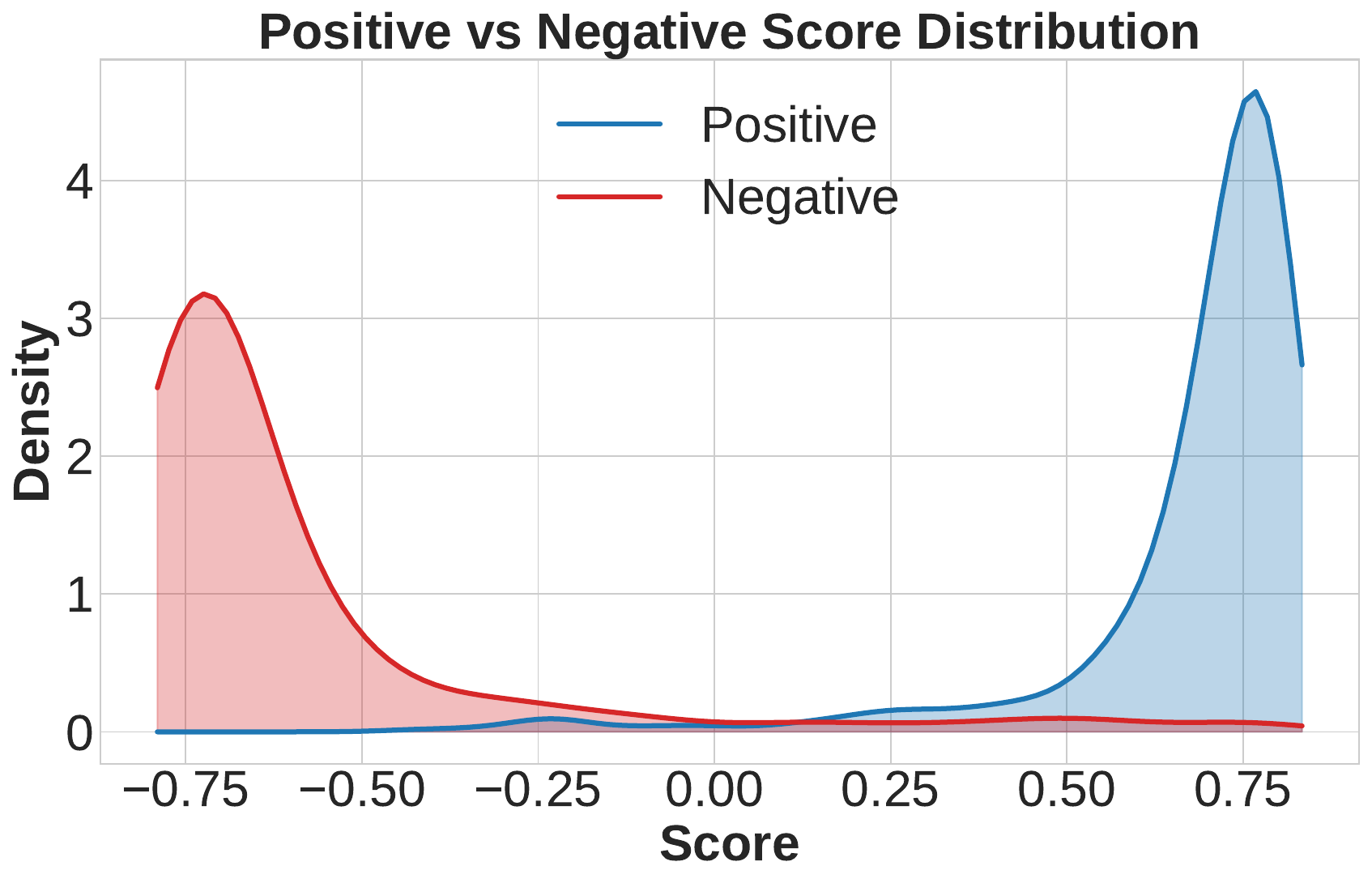}
        \label{fig:con_contra}}
        
    \end{minipage}
}
\caption{Visualization of contrastive scores}
\vspace{-0.2in}
\label{Model Choice}
\end{figure*}

\paragraph{PCA-Based Checking}

Principal Component Analysis (PCA) provides a powerful method for dimensionality reduction while preserving the most significant variations in data, making it particularly suitable for analyzing and differentiating LLMs' representation behaviors. Following the approach proposed by \citet{zou2023representation}, we first collect positive and negative sample pairs, then compute difference vectors for each pair. These difference vectors are calculated as: $ D_n = (-1)^n (v_n^+ - v_n^-)$, where \( v_n^+ \) and \( v_n^- \) are the internal representations of the positive and negative samples. The total number of pairs \( N \) is determined by the smaller sample size.

Next, we apply PCA to extract the top two principal components, \( P_1 \) and \( P_2 \), which define the subspace for analysis. All samples are then projected into this PCA space, reducing dimensionality while preserving variance. We assign binary labels to the projected samples: 1 for positive and 0 for negative. A logistic regression model is trained on this data to classify the two classes.

For new samples, we project their representations onto the PCA subspace and classify them using the trained logistic regression model.

\paragraph{Contrastive-learning-based checking.}
 Contrastive learning\cite{khosla2020supervised} offers an effective framework for differentiating complex data distributions by explicitly modeling relationships between positive and negative pairs. This approach highlights structural differences between samples, making it particularly suitable for tasks requiring nuanced behavioral distinctions. By maximizing the similarity among positive pairs while minimizing it for negative pairs, contrastive learning facilitates the extraction of discriminative features essential for classification. Consequently, we utilize contrastive learning to make the representations more distinguishable. The procedure is as follows:
\begin{enumerate}[topsep=0pt,itemsep=0pt,parsep=0pt,partopsep=0pt]

    \item Define a Contrastive Network: We design a contrastive network \(f_\theta: \mathbb{R}^d \rightarrow \mathbb{R}^h\) parameterized by \(\theta\), expressed as: $f_\theta(v) = \text{MLP}(v)$, where \(v\) represents the input vector among $V^+$ and $V^-$. The Multilayer Perceptron (MLP) serves as the backbone of our  network.

    \item Train the Network Using Contrastive Loss: We optimize the network using a contrastive loss function defined as:
\begin{equation} 
    \begin{aligned}
        \mathcal{L} = &\frac{1}{2} \left( \| f_\theta(v_i^+) - f_\theta(v_k^+) \|^2 \right) \\
        & + \max(0, m - \| f_\theta(v_i^+) - f_\theta(v_j^-) \|^2).
    \end{aligned}
\end{equation}
    where $k\in\{1,\cdots, N^{+}\}$, $m$ is the margin parameter that enforces a minimum distance between positive and negative samples. This formulation encourages the network to pull together similar positive samples while maintaining a separation from negative ones.

    \item Optimize the Network Parameters: The optimization problem is expressed as:
    \begin{equation}
        \theta^* = \argmin_\theta \mathbb{E}_{\{v_i\}, \{v_k^-\}, k} [\mathcal{L}]. \nonumber
    \end{equation}
    This step updates the parameters to minimize the contrastive loss, enhancing the model's ability to discern between positive and negative representations effectively.

    \item Compute Similarity Scores for Test Samples: For a test sample \(\tilde{v}\), we compute its similarity score with respect to the positive samples:
    \begin{equation}
        \text{score}(\tilde{v}) = \frac{1}{|V^+|} \sum_{v^+ \in V^+} s(f_{\theta^*}(\tilde{v}), f_{\theta^*}(v^+)), \nonumber
    \end{equation}
    where \(s(u, v)\) is the cosine similarity. This average similarity score serves as a measure of how closely the test sample aligns with the positive samples in the learned feature space.

 \item Classify the Test Sample: Finally, we classify the test sample based on a threshold \(t\):
    \begin{equation}
        \text{class}(\tilde{v}) = \begin{cases}
            \text{positive}, & \text{if } \text{score}(\tilde{v}) > t \\
            \text{negative}, & \text{otherwise} 
        \end{cases}\nonumber
    \end{equation}
\end{enumerate}
\subsection{Internal Knowledge Checking}
\label{Sec:query}


When presented with a query, it is crucial for LLMs to first assess whether they possess relevant internal knowledge. It can help the LLM determine whether to trigger retrieval and lays the foundation for subsequent checks, such as contradiction checking (Section \ref{Sec:conflict}). For our experimental dataset, we utilize the \href{https://github.com/hyintell/RetrievalQA}{RetrievalQA} dataset \cite{zhang2024retrievalqa}, a short-form open-domain question answering (QA) collection comprising 2,785 questions. This dataset includes 1,271 new world and long-tail questions that most LLMs cannot answer, serving as negative samples (queries without internal knowledge).  It also contains 1,514 questions that most LLMs can answer using only their internal  knowledge, functioning as positive samples (queries with internal knowledge). we randomly select 100 positive and 100 negative samples to anchor the PCA space, determine decision boundaries, and train the contrastive learning classifiers, and use the remaining data for evaluation. Mistral-7B-Instruct-v0.1 is used for this and following tasks.

We compare the representabtion-based methods with 2 types of \textbf{traditional checking} baselines, answer-based methods as well as probability-based methods. \textbf{Answer-based} methods mainly involves prompting LLMs and use their responses as checking results. We employ direct prompting as well as more sophisticated techniques such as In-Context Learning (ICL) and Chain-of-Thought (CoT) prompting to enhance the LLM's task comprehension. The prompting templates for each task are presented  in Appendix \ref{App:answer-prompts}, Table \ref{tab:Internal_Knowledge_Prompts}, Table \ref{tab:Context_Helpfulness_Prompts}, and Table \ref{tab:Internal_Belief_Alignment_Prompts}, respectively. We also employ \href{https://huggingface.co/SciPhi/SciPhi-Self-RAG-Mistral-7B-32k}{Self-RAG-Mistral}, a model fine-tuned to assess retrieval necessity and evidence relevance for tasks 1-3. It classifies by generating tokens like [retrieve] or [relevant]. See Appendix \ref{App:answer-prompts} for details. \textbf{Probability-based methods} involve analyzing the probabilities of LLMs' answers and comparing them with a threshold for classification. We employ three main indicators: overall perplexity as used by \citet{zou2024poisonedrag}, lowest probability score as implemented by \citet{jiang2023active}, and average probability score as utilized by \citet{wang2024self}. For each method, we vary the threshold and report the best accuracy while also plotting Receiver Operating Characteristic (ROC) curves and calculating the Area Under the Curve (AUC). Further details of these methods can be found in Appendix \ref{App: prob}.


\paragraph{Results.} We first evaluate whether answer-based methods or probability-based methods can handle internal knowledge checking. Table \ref{tab:rag_robustness} demonstrates that LLMs' own answers yield poor accuracy, even with advanced techniques like ICL and CoT. We observe high recall rates and numerous false-positive samples, suggesting LLMs' overconfidence in their knowledge and tendency to misclassify unknown queries as known.  The probability-based methods present relatively more promising results, achieving 69\% accuracy when using lowest scores. The ROC curves shown in Figure \ref{fig:roc_query} further illustrate this, with the lowest-scores method achieving the highest Area Under the Curve (AUC) of 0.74. This indicates that LLMs may exhibit lower confidence when encountering unknown queries. However, the overall accuracy is still far from reliable, indicating substantial room for improvement. For representation-based methods, we present performance results in Table \ref{tab:rag_robustness}, and provide visualizations of the PCA space and contrastive score distribution in  Figures \ref{fig:PCA_query} and \ref{fig:con_query}, respectively. As evidenced in Table \ref{tab:rag_robustness}, representation-based checking methods demonstrate significantly more promising results, with \textit{rep-PCA} achieving 75\% accuracy and \textit{rep-Con} reaching 79\% accuracy. Furthermore, Figures \ref{fig:PCA_query} and \ref{fig:con_query} clearly illustrate distinct distributions for queries with and without internal knowledge. These findings provide compelling evidence for the effectiveness of representation-based methods in internal knowledge checking.

\subsection{Uninformed Helpfulness Checking}
\label{Sec: Relevence_unknonw}
The retrieval process of RAG may return documents that are semantically related to the query but unhelpful in answering it. For example, "Einstein was born in Ulm, Germany in 1879 and later immigrated to the United States" is semantically related to the query "What year did Albert Einstein win the Nobel Prize in Physics?" but provides no answer. If an LLM lacks knowledge about the question, it's crucial to check whether the provided information actually helps answer the query, as the LLM can only use external knowledge to respond. In this subsection, we investigate whether LLMs' representations can perform well on such uninformed helpfulness checking tasks. To evaluate this, we use a subset of Natural Questions (NQ) \cite{kwiatkowski2019natural} employed by \citet{Cuconasu_2024}, containing 10,000 queries.\footnote{See Appendix \ref{App:dataset_checking} for knowledge checking datasets.} Each query in this dataset is associated with a golden passage (positive sample) that directly answers the question, as well as distractor passages retrieved from wikitext-2018  but not containing the answer. We use the distractor passage with the highest retrieval score as the negative sample. For uninformed helpfulness checking, we only use questions that Mistral-7B cannot correctly answer, totaling 8081 queries.  We randomly choose 100 positive and negative samples for the training of representation classifiers and use remaining data as test set. We also compared our methods with baselines as mentioned in Section \ref{Sec:query}.

\paragraph{Results.}
In Table \ref{tab:rag_robustness}, we present the performance of answer-based methods for helpfulness checking, as well as the best accuracy achieved by probability-based methods across various thresholds. We observe that although CoT (0.68) and Self-RAG (0.63) shows improved checking performance, the answer-based performance remains unsatisfactory and suffers from high false-positive rates. This indicates that LLM tends to regard unhelpful context as helpful in its responses. Furthermore, the accuracy of probability-based methods is also poor. We plot the  ROC curve  in Figure \ref{fig:roc_uninfor_help}, which shows low AUC values of 0.64 (Lowest Score), 0.62 (Average Score), 0.61 (Perplexity). This further indicates the differences in probability/perplexity between helpful and unhelpful contexts are not obvious and thus these matrics are not suitable for uninformed helpfulness checking.  In contrast, we can observe that representation-based methods demonstrate significantly better accuracy, with \textit{rep-PCA} achieving 79\% accuracy and \textit{rep-Contrastive} reaching 81\% accuracy, which is considerably more reliable. Figures \ref{fig:PCA_unknown_helpful} and \ref{fig:con_unknown_help} further illustrate that although some samples are difficult to distinguish and are misclassified, the majority of positive and negative pairs are distributed differently and can be effectively classified. These results clearly demonstrate the superiority of using representation-based methods for uninformed knowledge checking.

\subsection{Informed Helpfulness Checking}
\label{Sec:Relevence_known}
The integration of unhelpful documents may distract LLMs even when they possess internal knowledge about the question \cite{cuconasu2024power}. In this subsection, we evaluate whether the representation-based method can perform well for informed helpfulness checking. We utilize the same dataset and positive-negative pair settings as described in Section \ref{Sec: Relevence_unknonw}. However, for this evaluation, we select 1,919 queries that Mistral-7B can correctly answer, ensuring the model has internal knowledge about these queries. We randomly select 100 positive and negative samples to anchor the PCA space and train representation classifiers, while the remaining 1,819 positive-negative pairs are used for evaluation. We compares with same baselines mentioned in Section \ref{Sec:query}.
\paragraph{Results.}
The results of traditional checking methods are presented in Table \ref{tab:rag_robustness}. We observe that the performance of both answer-based and probability-based methods remains low for informed helpfulness checking. Furthermore, Figure \ref{fig:roc_infor_help} shows a low AUC value of of 0.60 (Lowest Score), 0.58 (Average Score), 0.59 (Perplexity).
These findings collectively indicate the limitations of these conventional methods in performing informed helpfulness checking effectively.
In contrast, Table \ref{tab:rag_robustness} demonstrates the superior performance of representation-based methods, with \textit{rep-PCA} achieving 81\% accuracy and \textit{rep-con} reaching 85\% accuracy. These results surpass those of uninformed helpfulness checking, possibly because the LLM's internal knowledge aids in better distinguishing between helpful and unhelpful sources. Figures \ref{fig:PCA_known_helpful} and \ref{fig:con_known_help} further illustrate that most positive and negative pairs are  distinguishable. These findings collectively demonstrate the success of representation-based methods in performing informed helpfulness checking.

\subsection{Contradiction Checking}
\label{Sec:conflict}

Previous research\cite{xieadaptive} has demonstrated that when presented with relevant but contradictory evidence, LLMs tend to prioritize external knowledge over their internal knowledge. Consequently, it is crucial to assess whether the provided external context aligns with or contradicts the LLM's internal beliefs. In this subsection, we investigate whether LLMs' representations can serve as more reliable indicators of contradictions between external context and the model's internal knowledge. we utilize a subset of \href{https://github.com/OSU-NLP-Group/LLM-Knowledge-Conflict/tree/main}{ConflictQA} \cite{xieadaptive}. Each sample contains a PopQA\cite{mallen-etal-2023-trust} question, correct aligned evidence, and ChatGPT-generated contradictory evidence. See appendix \ref{App:dataset_checking} for details. We sampled 1146 questions that Mistral-7B answers correctly, using aligned evidence with the query as positive samples and contradictory evidence as negative samples.
 We utilized 10\% of the dataset (114 positive-negative pairs) to anchor the PCA space, calculate decision boundaries, and train the contrastive learning classifiers. The remaining 90\% was reserved for testing purposes. We compare representation based method with traditional methods in  Section \ref{Sec:query}.



\paragraph{Results.}

We initially assess whether LLMs' answers and their associated probability/perplexity metrics can effectively indicate contradictions. The results in Table \ref{tab:rag_robustness} reveal that LLMs' answers continue to exhibit low accuracy and suffer from a high rate of false positives. This suggests that LLMs tend to interpret contradictory external knowledge as aligned evidence in their responses. Furthermore, Figure \ref{fig:roc_contra} demonstrates a extremenly low AUC of of 0.39 (Lowest Score), 0.34 (Average Score), 0.33 (Perplexity), indicating minimal differences in probability/perplexity distributions when LLMs are presented with aligned versus contradictory evidence.
As illustrated in Table \ref{tab:rag_robustness}, representation-based methods demonstrate significantly superior performance, with \textit{rep-PCA} achieving 91\% accuracy and \textit{rep-Contrastive} attaining an impressive 95\% accuracy. Our visualizations, presented in Figures \ref{fig:PCA_contra} and \ref{fig:con_contra}, reveal distinct distributions and contradictory scores for the contradictory and aligned contexts. These pronounced differences strongly indicate that our method can effectively discriminate between these context types.


%% file: Sections/Experiment.tex
\section{Representation Based Context Filtering}
In this section,  we investigate how knowledge checking based on representations affect performance of RAG systems. 

\subsection{Representation Based Filtering}
\label{Sec: filter_methods}
 We design a simple representation-based context filtering strategy.  We perform representation checking on our test queries and retrieved documents. First, we conduct internal knowledge checking to identify known and unknown queries. Next, we apply helpfulness checking to all queries and contradictory checking only to predicted known queries. Finally, we filter out contexts classified as unhelpful or contradictory. We incorporate such filtering with Mistral-7B-v0.1, Llama-2-7B-Chat as well as Llama-3-8B-Instruct. The classifiers for knowledge checking are trained using datasets from Sections \ref{Sec:query}, \ref{Sec: Relevence_unknonw}, \ref{Sec:Relevence_known}, and \ref{Sec:conflict} respectively \footnote{We still refer to our methods as \textit{Rep-PCA} and \textit{Rep-Con} based on which knowldege checking methods we use.}.

\subsection{Experiment Setup}
\label{Sec:filter_set_up}
\begin{table*}[htbp]
\centering
\caption{Overall results on NQ and PopQA }
\vspace{-0.1in}
\label{tab:overall_results}
\resizebox{\textwidth}{!}{%
\begin{tabular}{llcccc}
\hline
\multirow{2}{*}{Retrieval Type} & \multirow{2}{*}{Model} & \multicolumn{2}{c}{NQ} & \multicolumn{2}{c}{PopQA} \\
\cline{3-6}
 &  & Noisy Acc (\%) & Clean Acc(\%) & Noisy Acc (\%) & Clean Acc(\%) \\
\hline
\multirow{5}{*}{No-retrieval} 
 & LLaMA\textsubscript{2-7B-Chat}\cite{touvron2023llama} & 73.17\% & 29.03\% & 71.20\% & 19.60\% \\
 & LLaMA\textsubscript{3-8B-Instruct}\cite{llama3modelcard} & 80.86\% & 32.73\% & 74.16\% & 22.45\% \\
 & Mistral\textsubscript{7B-Instruct}\cite{jiang2023mistral} & 97.21\% & 20.10\% & 98.02\% & 15.58\% \\
 & Alpaca\textsubscript{7B}\cite{alpaca} & 72.61\% & 23.94\% & 71.84\% & 13.07\% \\
 & Vicuna\textsubscript{7B}\cite{zheng2023judging} & 73.16\% & 26.64\% & 74.56\% & 19.43\% \\
\hline
\multirow{5}{*}{Unfiltered} 
 & LLaMA\textsubscript{2-7B-chat} & 34.66\% & 26.96\% & 60.91\% & 45.90\% \\
 & LLaMA\textsubscript{3-8B-Instruct} & 48.12\% & 33.59\% & 51.27\% & 40.54\% \\
 & Mistral\textsubscript{7B-instruct} & 28.97\% & 24.35\% & 55.96\% & 48.58\% \\
 & Alpaca\textsubscript{7B} & 37.12\% & 29.80\% & 62.65\% & 53.10\% \\
 & Vicuna\textsubscript{7B} & 36.12\% & 28.28\% & 54.35\% & 49.75\% \\
\hline
\multirow{12}{*}{Filtered} 
 & Direct filtering & 30.08\% & 24.32\% &54.05\%  & 46.31\% \\
 & ICL filtering & 29.90\% & 23.95\% & 55.28\% & 47.02\% \\
 & CoT filtering & 30.19\% & 24.18\% & 56.03\% & 46.95\% \\
 & Self-RAG\textsubscript{Llama-2} &39.10\% & 30.27\% & 65.17\% & 52.08\% \\
 & Self-RAG\textsubscript{Mistral} &  32.30\%& 26.07\% & 60.65\% & 50.57\% \\
 & \textit{Rep-PCA(Mistral)} & 70.73\% & 29.81\% & 73.63\% & 56.16\% \\
 & \textit{Rep-Con(Mistral)} & 72.53\% & 32.39\% & 72.62\% & 57.62\% \\
 & \textit{Rep-PCA(Llama-2)} & 67.93\% & 31.32\% & 66.78\% & 53.97\% \\
 & \textit{Rep-Con(Llama-2)} & 69.95\% & 33.64\% & 67.59\% & 54.26\% \\
& \textit{Rep-PCA(Llama-3)} & 67.81\% & 35.32\% & 71.16\% & 50.18\% \\
 & \textit{Rep-Con(Llama-3)} & 69.81\% & 36.75\% & 72.16\% & 52.26\% \\
\hline
\end{tabular}%
}
\vspace{-0.1in}
\end{table*}

\paragraph{Datasets.} For our evaluation, we utilize two primary datasets: a subset of Natural Questions (NQ) used by \citet{Cuconasu_2024}, comprising 83,104 queries with gold documents of 512 tokens or less, and ConflictQA, a subset of PopQA containing 11,216 queries with labeled golden passages and misleading contexts, as employed by \cite{xieadaptive}. We use Wikipedia-2018 as  retrieval database, injecting golden passages for queries not already present. To assess RAG performance in the presence of misleading information, we further categorize the queries into "noisy" and "clean" sets. For noisy queries, we selected 1,000 from NQ and 500 from PopQA that Mistral-7B can correctly answer and other LLMs we use can achieve over 70\% accuracy on. The remaining queries are categorized as clean. We injected misleading contexts of those noisy queries to retrieval DB. For ConflictQA, we used the misleading contexts provided by \citet{xieadaptive}. For NQ, we constructed them using ChatGPT. \footnote{Details of datasets are available in Appendix \ref{App:dataset_filtering}.}


\paragraph{RAG pipeline.} 
Our retrieval database comprises the corpus from Wikipedia-2018 following~\citet{jiang2023active}, as well as misleading passages for noisy queries. Each document in the wiki-text-2018 is segmented into non-overlapping passages of 100 words. Each misleading passage is kept whole without further segmentation. We utilize Contriever \cite{izacard2021unsupervised} to construct the embeddings of the retrieval dataset and index them using FAISS \cite{douze2024faiss}, following the settings outlined by \citet{cuconasu2024power}. We begin by retrieving the top-10 documents from the database. For baselines without filtering, we directly select the top-2 documents with the highest retrieval scores as contexts. For methods with filtering, we choose top-2 unfiltered documents with the highest retrieval scores.


\paragraph{Baselines.} We compare represntation-based methods against various baselines, including no-retrieval and retrieval w/o filtering predictions with different models (Mistral-7B-v0.1, Llama-2-7B-Chat, Llama-3-8B-Instruct, Vicuna-7B, and Alpaca-7B), and traditional filtering methods. For Direct, ICL, and CoT filtering, we perform answer-based knowledge checking as described in Sections \ref{Sec:query}. We then filter out unhelpful contexts, and contradictory contexts for predicted known queries. We only filter out irrelevant contexts for Self-RAG, as it does not provide contradiction checking.

\paragraph{ Metrics.} We report the exact match accuracy for clean (Clean Acc) and noisy queries (Noisy Acc).

\begin{figure}[t]
\centering
\begin{tabular}{@{}cc@{}}
    \includegraphics[width=0.49\columnwidth]{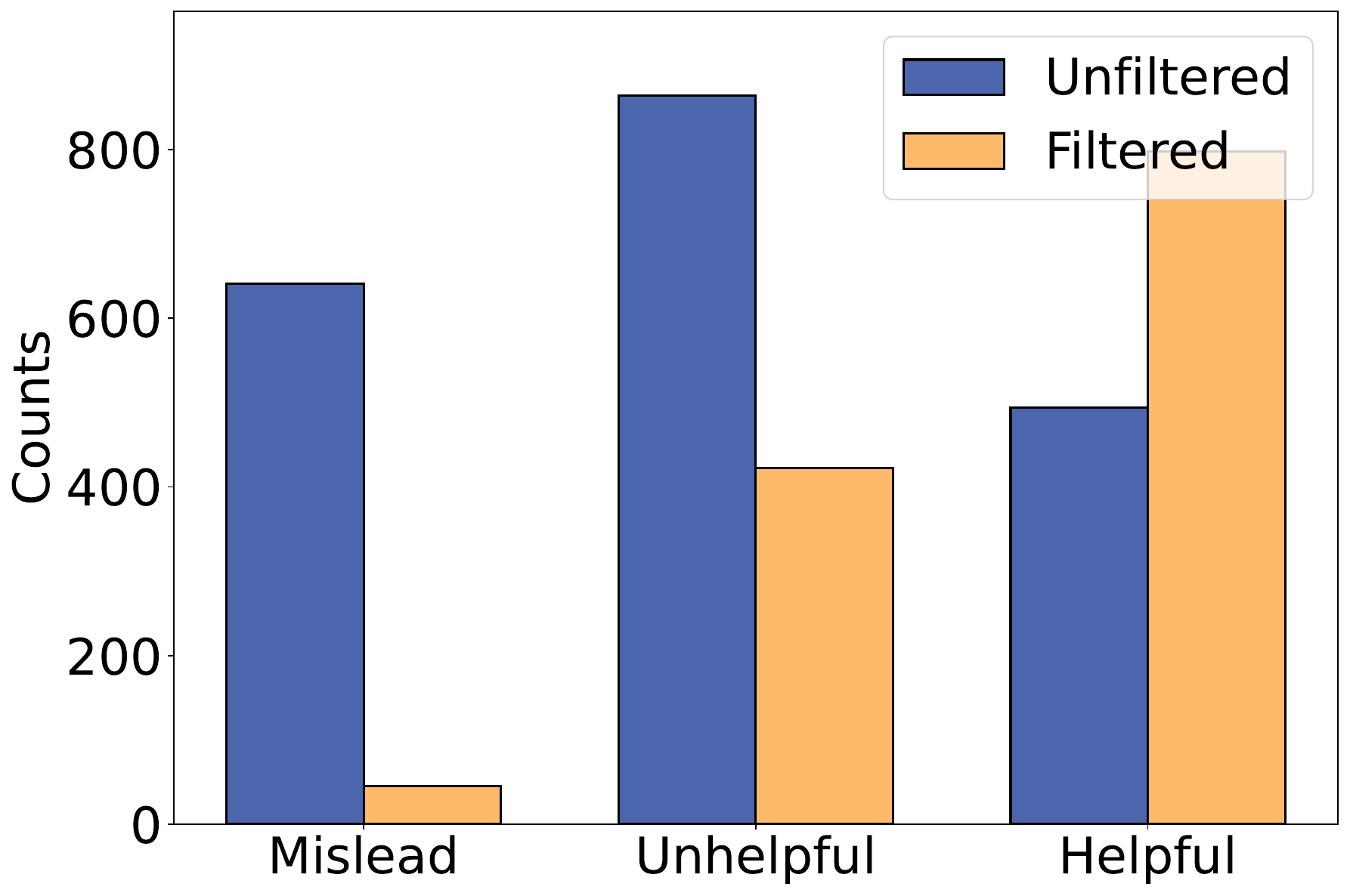} &
    \includegraphics[width=0.49\columnwidth]{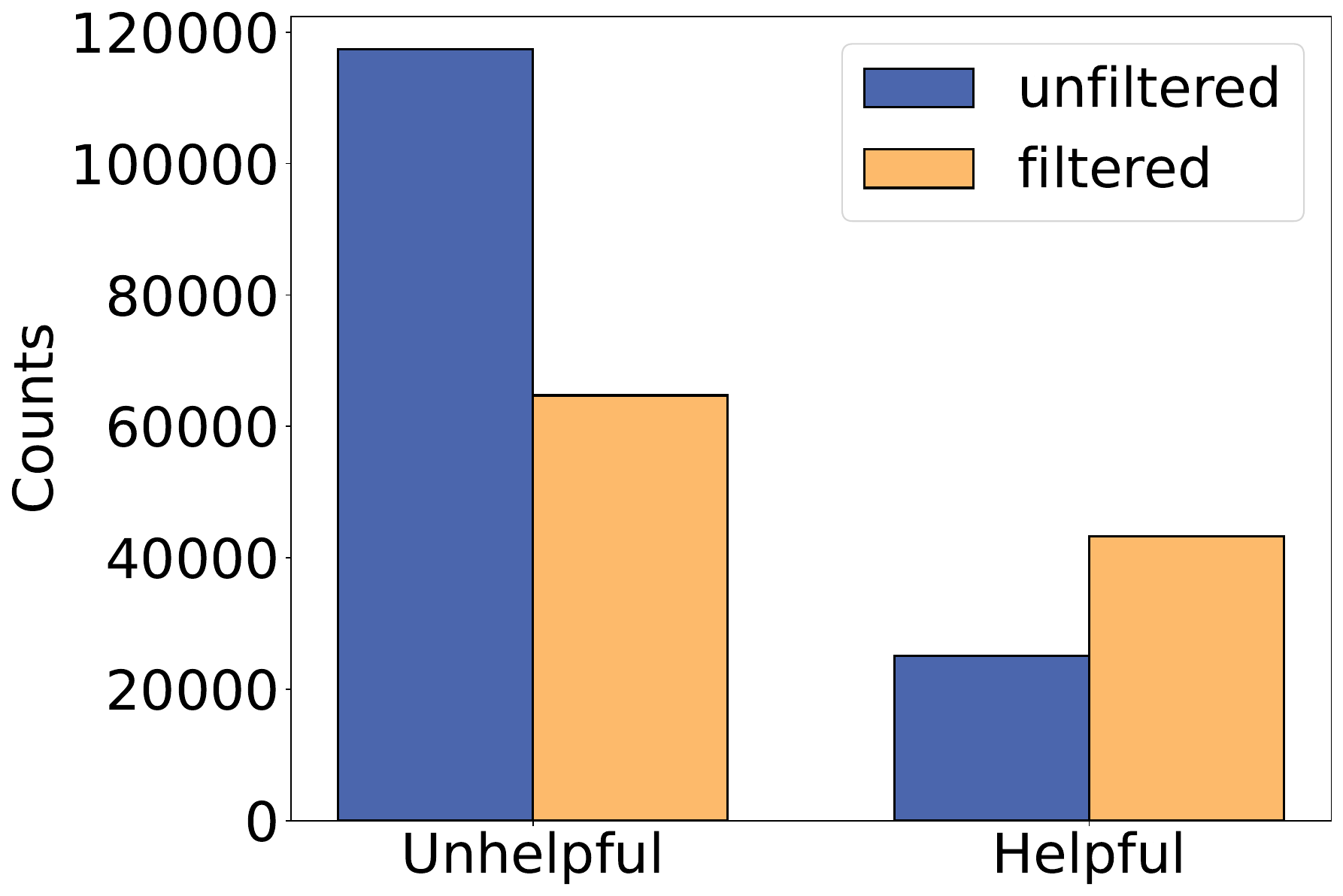} \\
    (a) Noisy queries & (b) Clean queries
\end{tabular}
\vspace{-0.1in}
\caption{Filtering results}
\vspace{-0.3in}
\label{fig:filter}
\end{figure}
\subsection{Performance on Clean Queries} 
\label{Sec:results_clean}

The results in Table \ref{tab:overall_results} demonstrate that our method achieves better Clean Acc(\%) compared to unfiltered baselines. For instance, \textit{Rep-Con(Mistral)} shows an 8.04\% increase in accuracy on NQ and an 8.84\% increase on PopQA compared to retrieval without filtering. This improvement indicates that representation methods can effectively filter out unhelpful contexts and subsequently enhance RAG performance. In contrast, other filtering baselines show minimal improvement over no filtering, aligning with our findings in Section \ref{Sec: Representation} that they have limitations in effective knowledge checking.

\subsection{Performance on Noisy Queries}
\label{Sec:results_noisy}
The results in Table \ref{tab:overall_results} reveal that injecting misleading contexts significantly impairs LLMs' performance on noisy queries. For instance, Mistral-7B's performance on NQ noisy queries drops by more than 70\% compared to zero-shot generation. However, our filtering mechanism effectively mitigates this issue, even when misleading contexts are retrieved. Notably, on noisy NQ queries, \textit{Pre-con(Mistral)} recovers the noisy accuracy from 28.97\% to 72.53\%, a substantial 43.56\% improvement. Similarly, on noisy PopQA queries, it recovers accuracy from 55.96\% to 73.64\%. Furthermore, representation-based filtering consistently outperforms other filtered baselines, validating its effectiveness in filtering out misleading knowledge. These results indicating that representation-based filtering can boost RAG systems' robustness against noisy contexts.

\subsection{Documents Distribution after Filtering}
\label{Sec:results_dis}
In this subsection, we analyze the distribution of  unhelpful, misleading, and helpful documents used as contexts before and after our filtering process\footnote{We categorize injected misleading contesxts as "misleading", contexts with right answers as "helpful" otherwise "unhelpful".}. Figure \ref{fig:filter} shows the results for both noisy and clean queries from the NQ dataset\footnote{See Appendix~\ref{app:popqa_filter} for PopQA results.}. For noisy queries, our filtering method demonstrates remarkable effectiveness by almost entirely eliminating misleading contexts and significantly reducing unhelpful ones. Consequently, the number of helpful contexts increases, as some unhelpful and misleading contexts with high retrieval scores are filtered out. Similarly, for clean queries, we observe a decrease in unhelpful documents and an increase in helpful ones. These results validate the effectiveness of our representation-based  checking. The improved context quality from this filtering process is the key reason for the performance increase.

%% file: Sections/Conclusions.tex
\vspace{-0.1in}
\section{Conclusions}
\label{Conclusion}
This study delves into the knowledge checking in RAG systems. To achieve this goal, we identified and proposed four key tasks. Through comprehensive analysis of LLMs' representation behaviors, we found that representation-based methods significantly outperform answer-based or probability-based approaches. Leveraging these findings, we developed representation-based classifiers for knowledge filtering. Results demonstrate that simply filtering of contradictory and unhelpful knowledge substantially improves RAG performance.




%% file: Sections/Limitations.tex
\section{Limitations}
In this work, we have demonstrated that the representations of LLMs can significantly enhance the robustness of RAG systems. However, the underlying mechanisms by which LLMs identify, utilize, and integrate external knowledge with their internal knowledge remain an open research question. Our framework employs \textit{Rep-PCA} and introduces \textit{Rep-Contra} for context analysis. While these methods have shown promising results, we aim to explore more sophisticated analytical approaches. It is important to note that a significant challenge lies beyond the scope of our current work: determining the correctness of context when the LLM itself lacks knowledge about the question at hand. This presents a more complex problem, and we posit that external sources may be necessary, as LLMs' self-signals alone may not be sufficient to fully address this challenge.

%% file: Sections/Appendix.tex
\clearpage

\appendix
\onecolumn
\section{Appendix}
\href{}{}
\subsection{Ablation Studies}
\label{app:ablation}
\subsubsection{Using Other Layers' Representation}
\label{app:other_layer}
\begin{figure*}[htpb]
\centering
\resizebox{\textwidth}{!}{%
    \begin{minipage}{\textwidth}
        \subfloat[Internal Knowledge ]{\includegraphics[width=.25\textwidth]{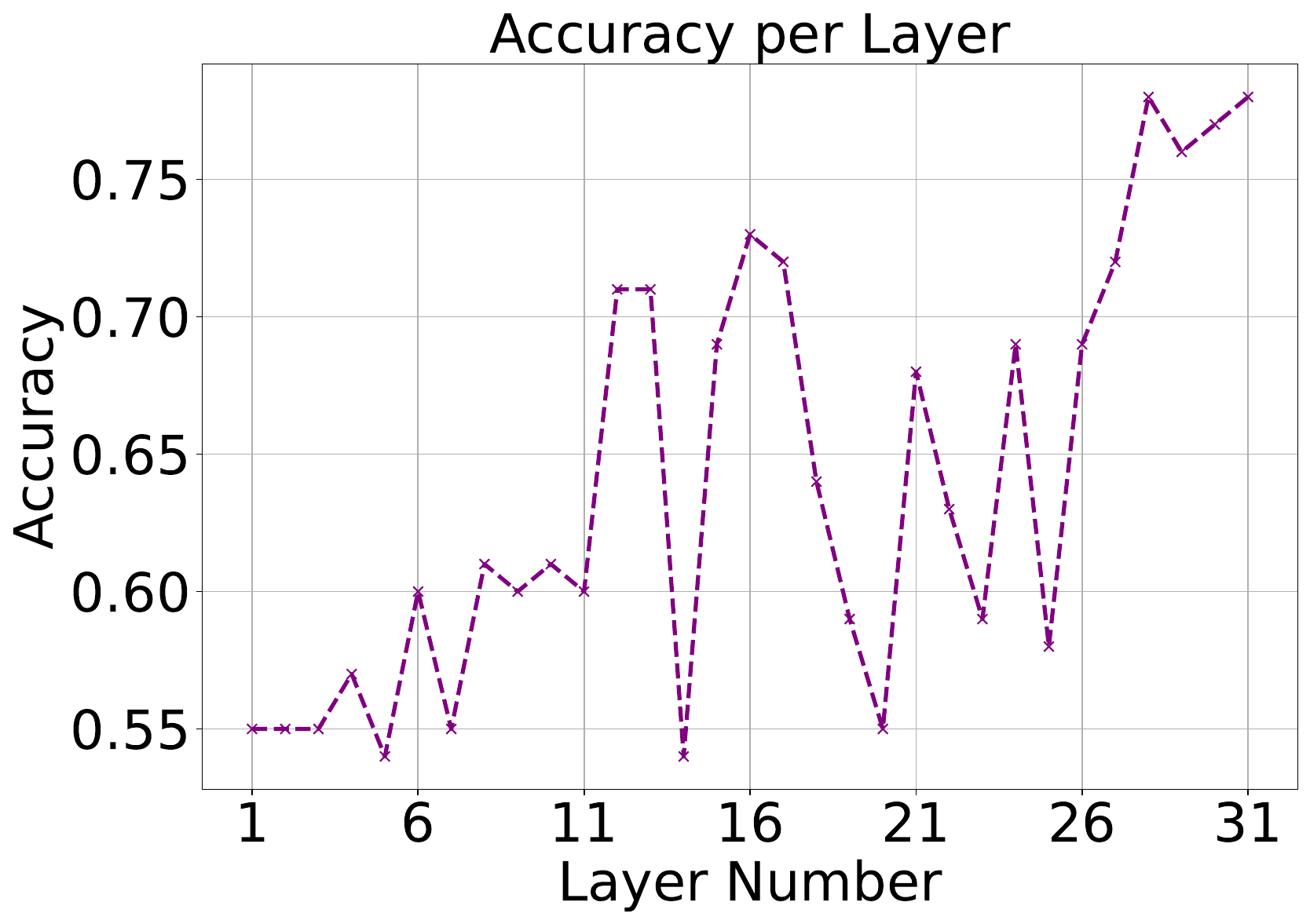}
        \label{fig:query_layer}}
        \subfloat[Uninformed helpfulness]{\includegraphics[width=.25\textwidth]{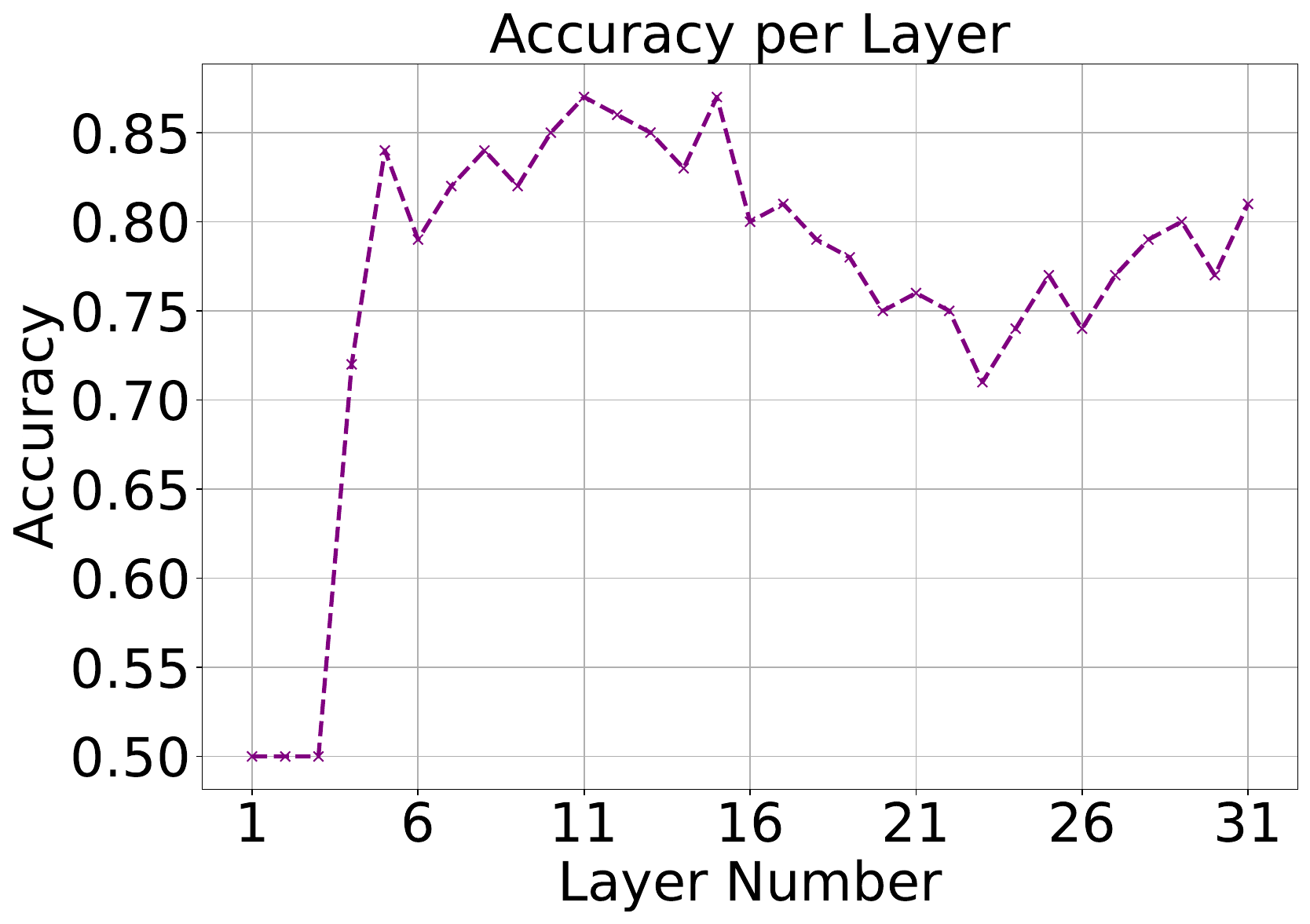}
        \label{fig:unknown_helpful_layer}}
        \subfloat[Informed helpfulness]{\includegraphics[width=.25\textwidth]{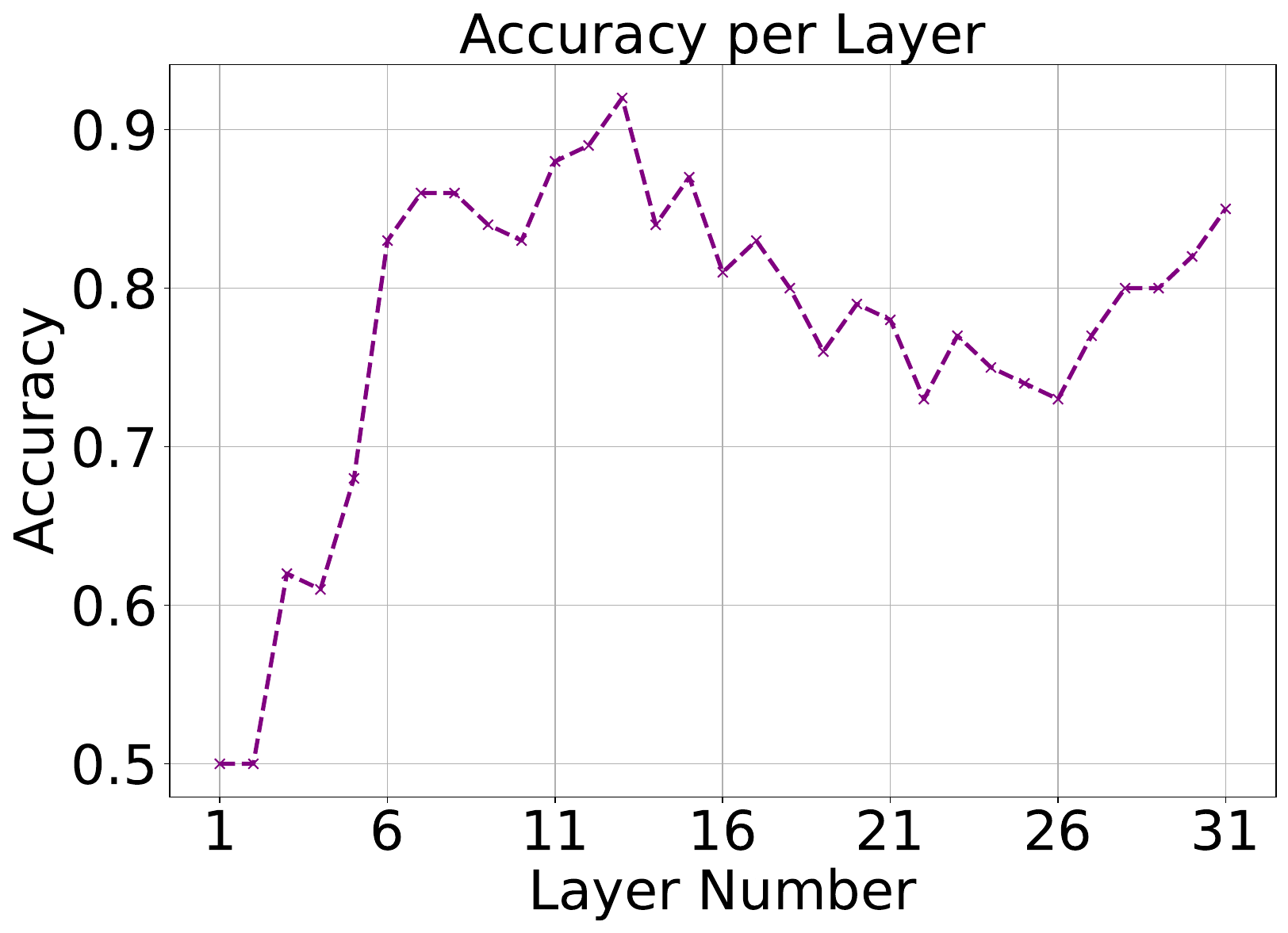}
        \label{fig:known_helpful_layer}}
        \subfloat[Contradiction]{\includegraphics[width=.25\textwidth]{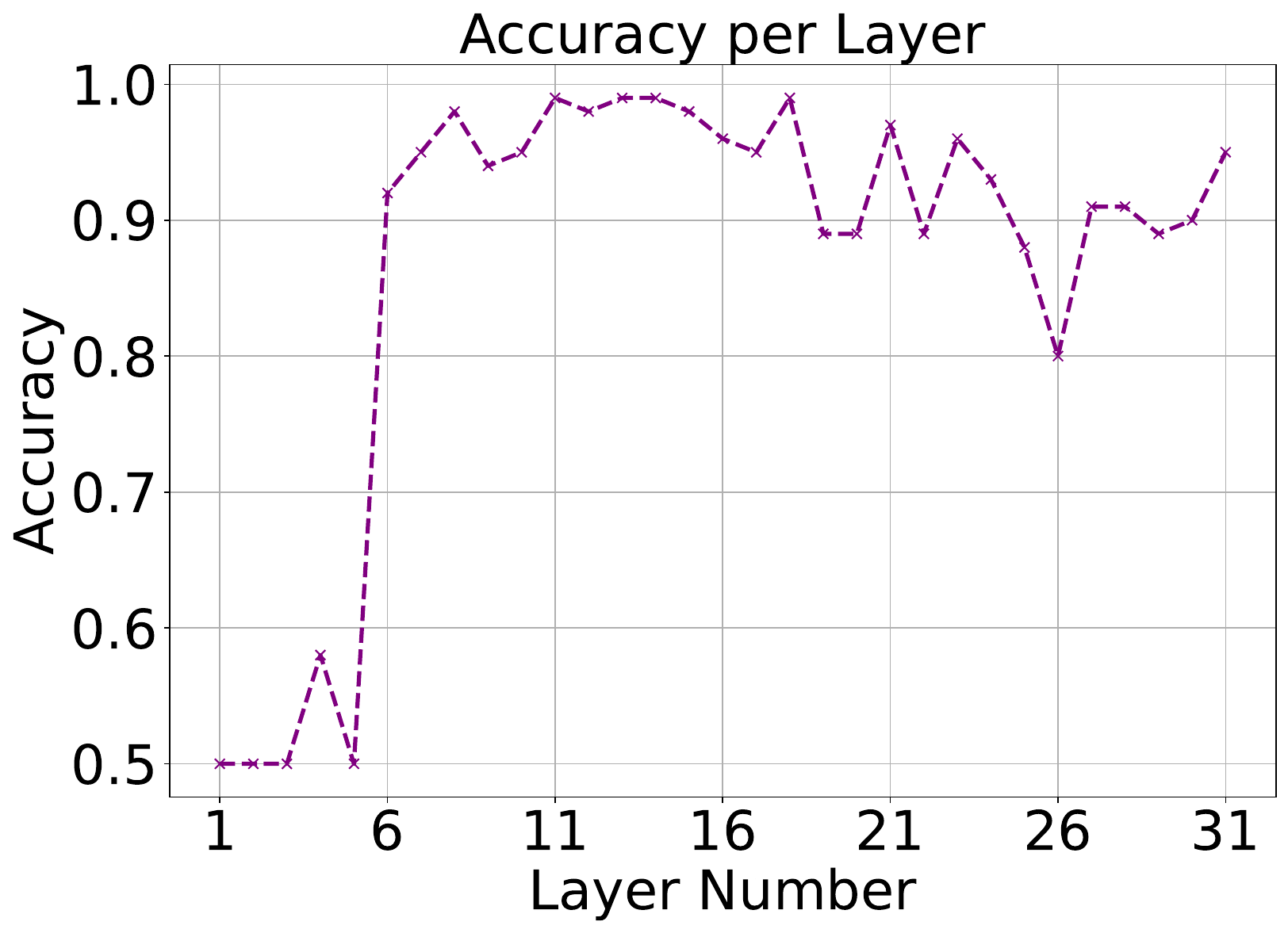}
        \label{fig:contra_layer}}
        
    \end{minipage}
}
\caption{Accuracy on different layers}
\label{performance_layer}
\end{figure*}

In our main section, we primarily base our analysis on the representations from the last layers. We also explore the knowledge checking performance using representations from other layers. Figure \ref{performance_layer} illustrates the 'rep-con' performance of each layer across four different tasks.
We observe that the performance using the first few layers is poor for all tasks. This may be because these layers primarily capture low-level features and patterns in the input, rather than higher-level semantic concepts. They haven't yet integrated this information into more abstract or task-relevant representations, which are necessary for complex knowledge checking tasks. For internal knowledge checking tasks, using representations from the last few layers shows the best performance. However, for other tasks, representations from some middle layers perform better than those from the last layer. This may be because these middle layers are more responsible for processing corresponding concepts. In practice, we suggest using a validation set to identify the layers with the best performance and using the results from these layers for knowledge checking.

\begin{figure*}[htpb]
\centering
\resizebox{\textwidth}{!}{%
    \begin{minipage}{\textwidth}
        \subfloat[Internal Knowledge ]{\includegraphics[width=.25\textwidth]{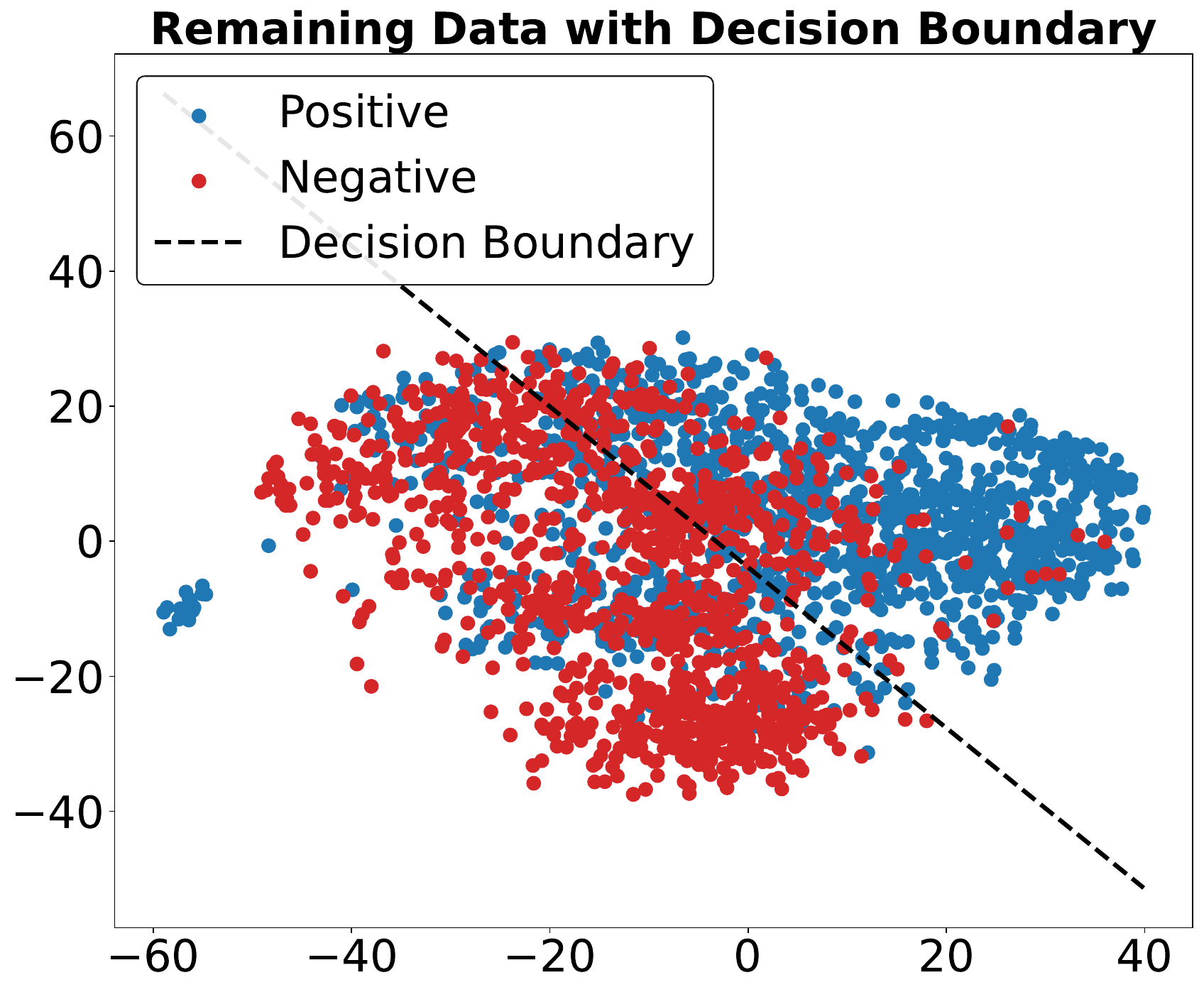}
        \label{fig:PCA_query_llama}}
        \subfloat[Uninformed helpfulness]{\includegraphics[width=.25\textwidth]{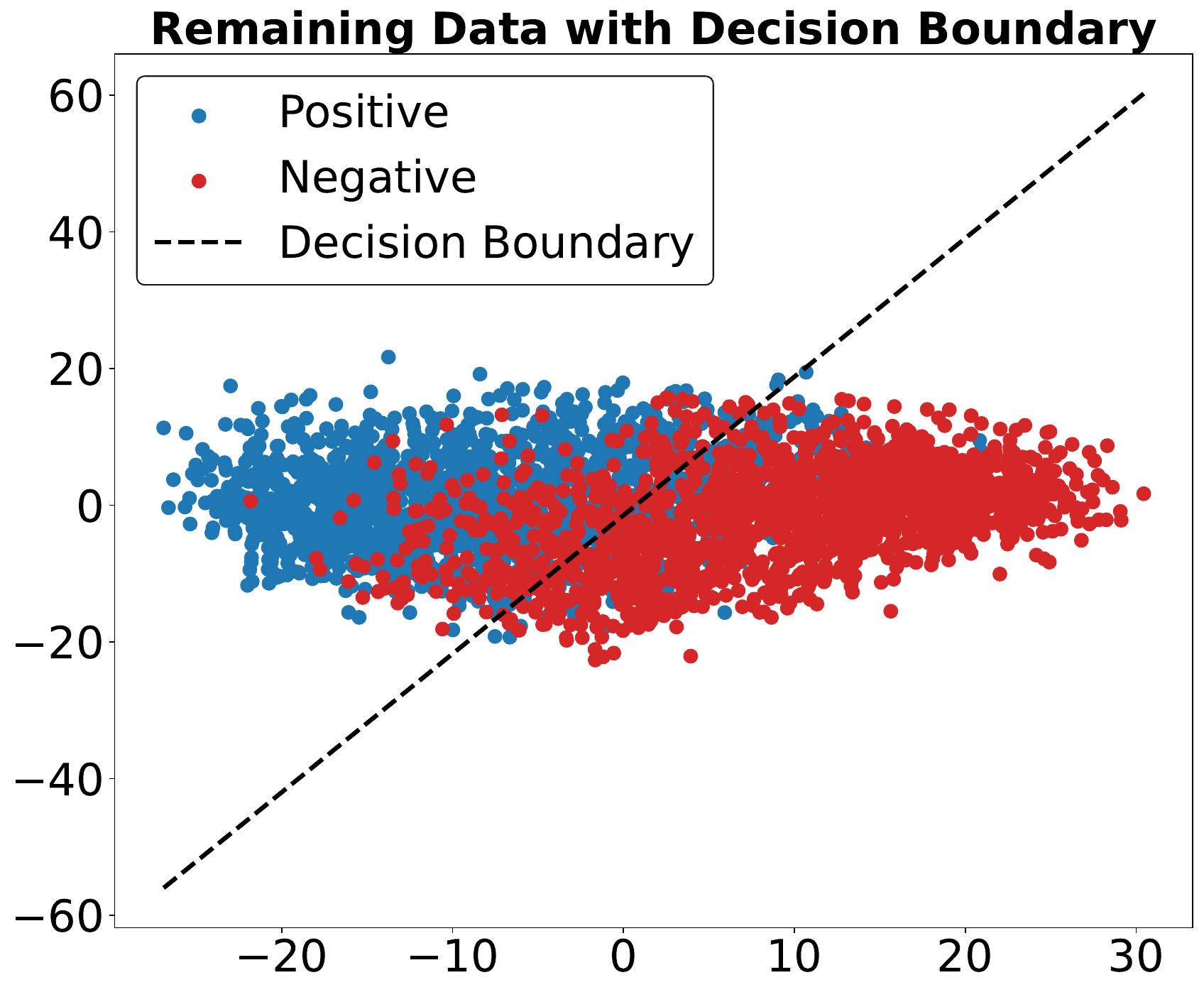}
        \label{fig:PCA_unknown_helpful_llama}}
        \subfloat[Informed helpfulness]{\includegraphics[width=.25\textwidth]{pic/Remaining_decision_boundary_distract_poison-Llama-2-7B.pdf}
        \label{fig:PCA_known_helpful_llama}}
        \subfloat[Contradiction]{\includegraphics[width=.25\textwidth]{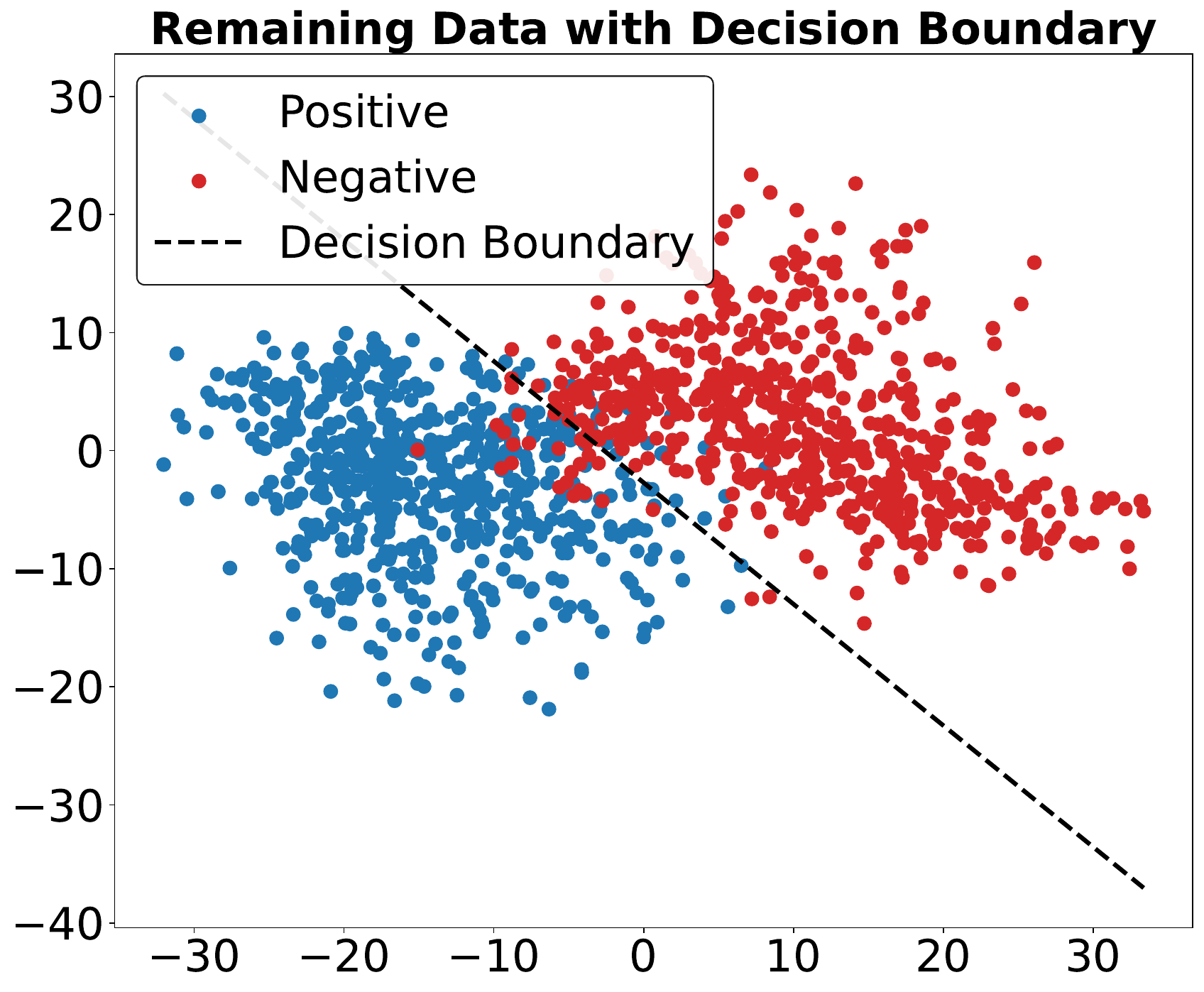}
        \label{fig:PCA_contra_llama}}
        
    \end{minipage}
}
\caption{Visualization on PCA space(Llama-2-7B-Chat)}
\label{PCA_llama2}
\end{figure*}

\begin{figure*}[htbp]
\centering
\resizebox{\textwidth}{!}{%
    \begin{minipage}{\textwidth}
        \subfloat[Internal Knowledge ]{\includegraphics[width=.25\textwidth]{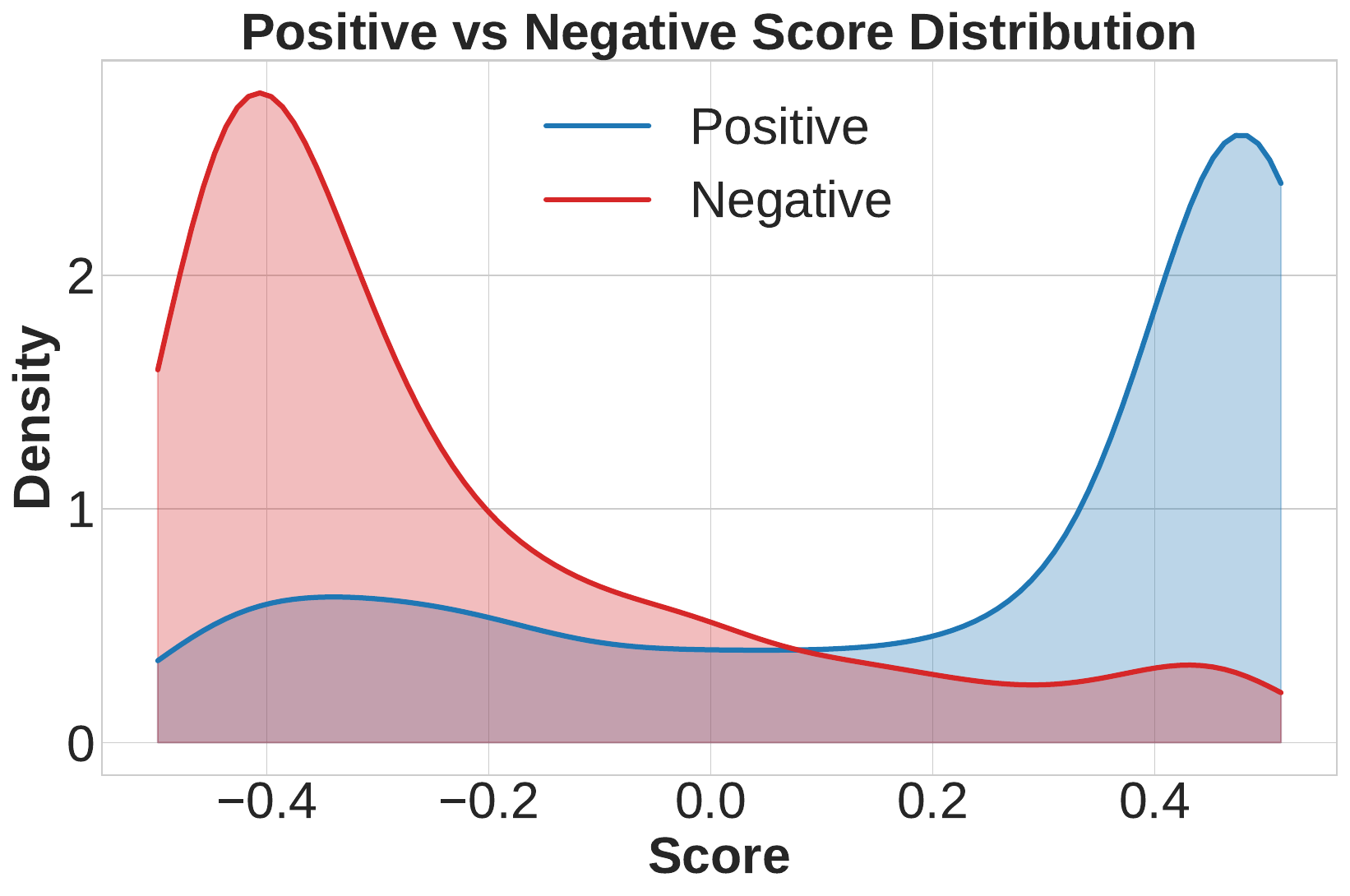}
        \label{fig:con_query_llama}}
        \subfloat[Uninformed helpfulness]{\includegraphics[width=.25\textwidth]{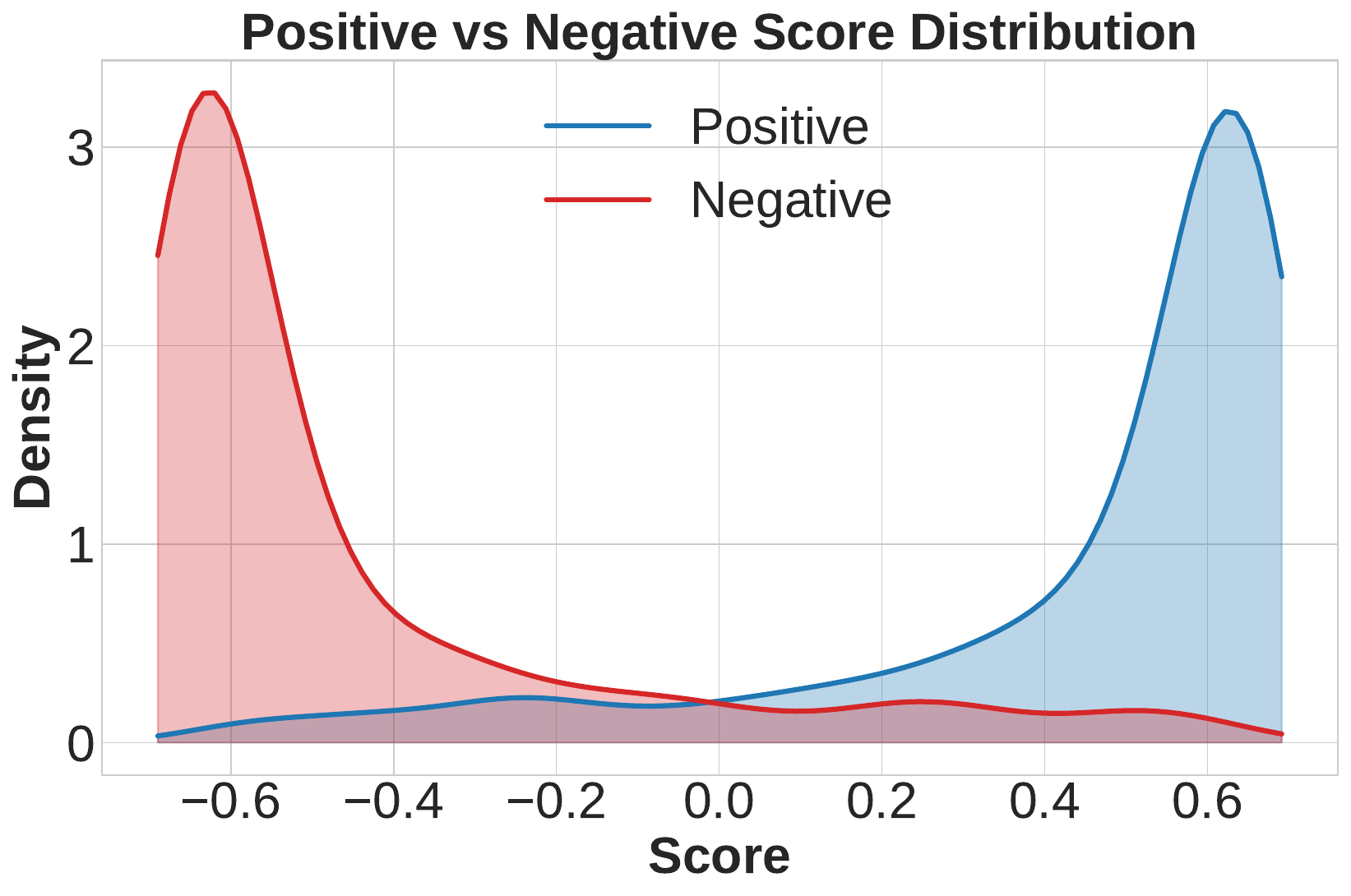}
        \label{fig:con_unknown_help_llama}}
        \subfloat[Informed helpfulness]{\includegraphics[width=.25\textwidth]{pic/Remain_contrastive_distract_poison-Llama-2-7B-all.pdf}
        \label{fig:con_known_help_llama}}
        \subfloat[Contradiction]{\includegraphics[width=.25\textwidth]{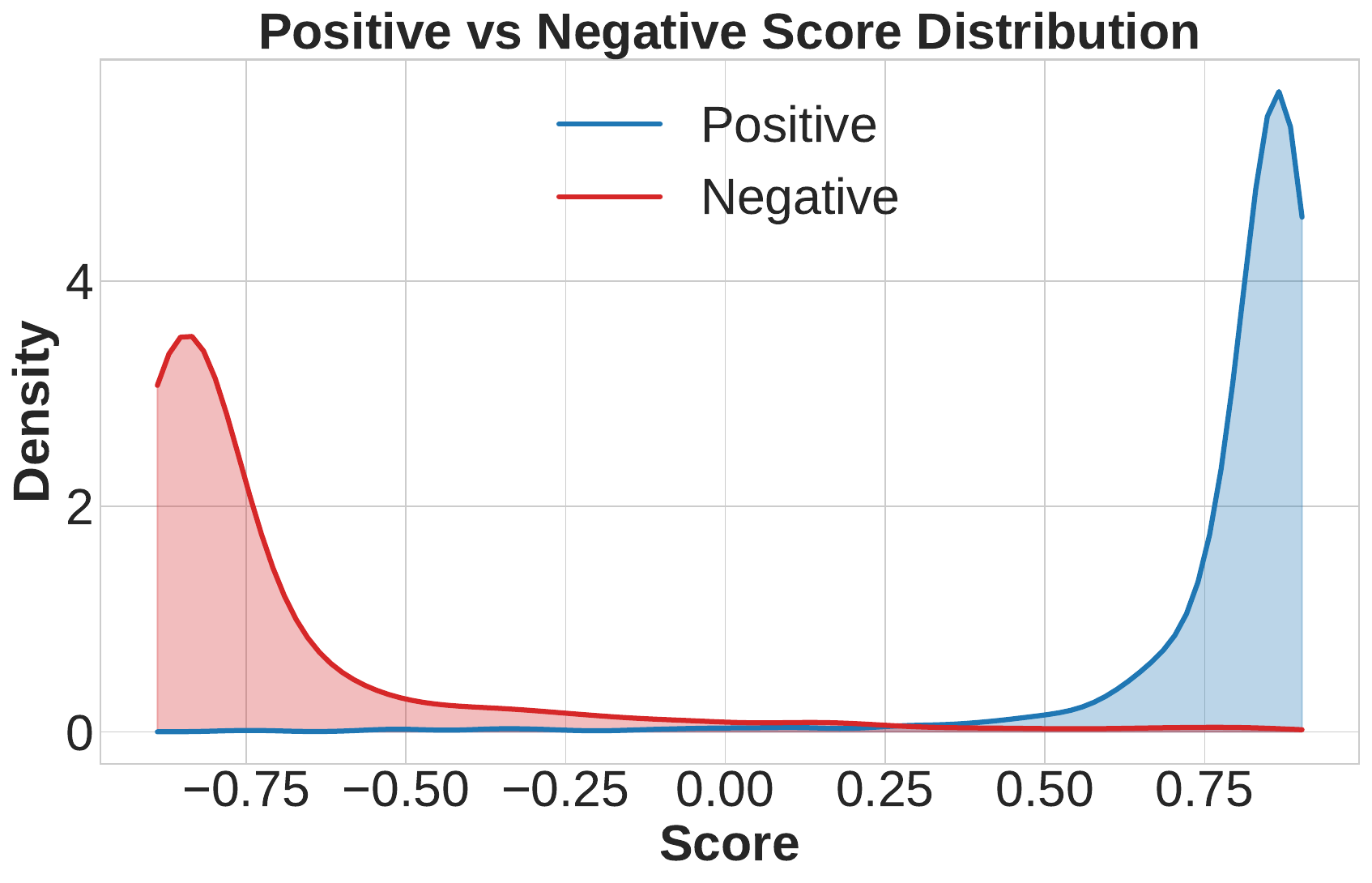}
        \label{fig:con_contra_llama}}
        
    \end{minipage}
}
\caption{Visualization of contrastive scores(Llama-2-7B-Chat)}
\label{Contrastive_scores_llama-2}
\end{figure*}

\begin{table*}[!htpb]
\centering
\caption{Representation checking performance of Llama-2-7B }
\label{tab:rag_robustness_llama2}
\resizebox{\textwidth}{!}{
\begin{tabular}{@{}l|cccc|cccc|cccc|cccc@{}}
\toprule
 & \multicolumn{4}{c|}{Internal Knowledge} & \multicolumn{4}{c|}{Uninformed Helpfulness} & \multicolumn{4}{c|}{Informed Helpfulness} & \multicolumn{4}{c}{Conflict Detection} \\
\cmidrule(l){2-5} \cmidrule(l){6-9} \cmidrule(l){10-13} \cmidrule(l){14-17}
Method & Acc & Pre & Rec & F1 & Acc & Pre & Rec & F1 & Acc & Pre & Rec & F1 & Acc & Pre & Rec & F1 \\
\midrule
Re-PCA & 0.75 & 0.80 & 0.73 & 0.76 & 0.79 & 0.77 & 0.81 & 0.79 & 0.83 & 0.84 & 0.82 & 0.83 & 0.96 & 0.96 & 0.96 & 0.96 \\
Re-Contra & 0.76 & 0.83 & 0.72 & 0.78 & 0.81 & 0.80 & 0.82 & 0.81 & 0.89 & 0.89 & 0.89 & 0.89 & 0.97 & 0.96 & 0.99 & 0.97 \\
\bottomrule
\end{tabular}
}
\end{table*}
\subsubsection{Knowledge Checking Performance of Other model}
\label{app:other_model}
In this section, we also visulize and report the representation knowledge checking performance of Llama2-7B-Chat model. From the results of Table \ref{tab:rag_robustness_llama2} and visualization in Figure \ref{PCA_llama2} and Figure \ref{Contrastive_scores_llama-2}, we can get similar observation as Mistral-7B, the performance of representation-based checking is also promising for 4 tasks. Indicating the generalizbility of representation knowlege checking across models.
\subsubsection{Filtering Results on PopQA}
\label{app:popqa_filter}
\begin{figure}[t]
\centering
\begin{tabular}{@{}cc@{}}
    \includegraphics[width=0.3\columnwidth]{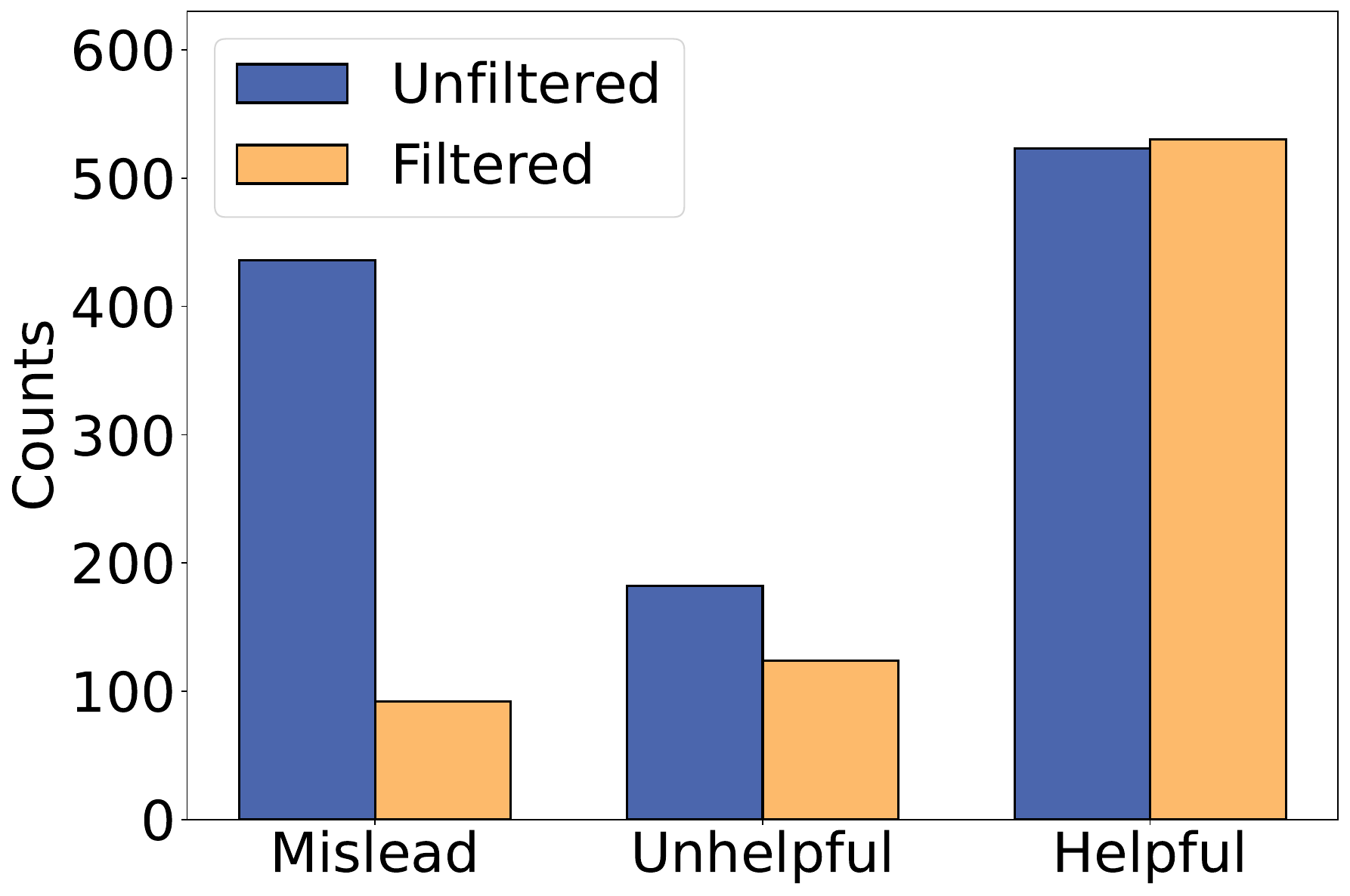} &
    \includegraphics[width=0.3\columnwidth]{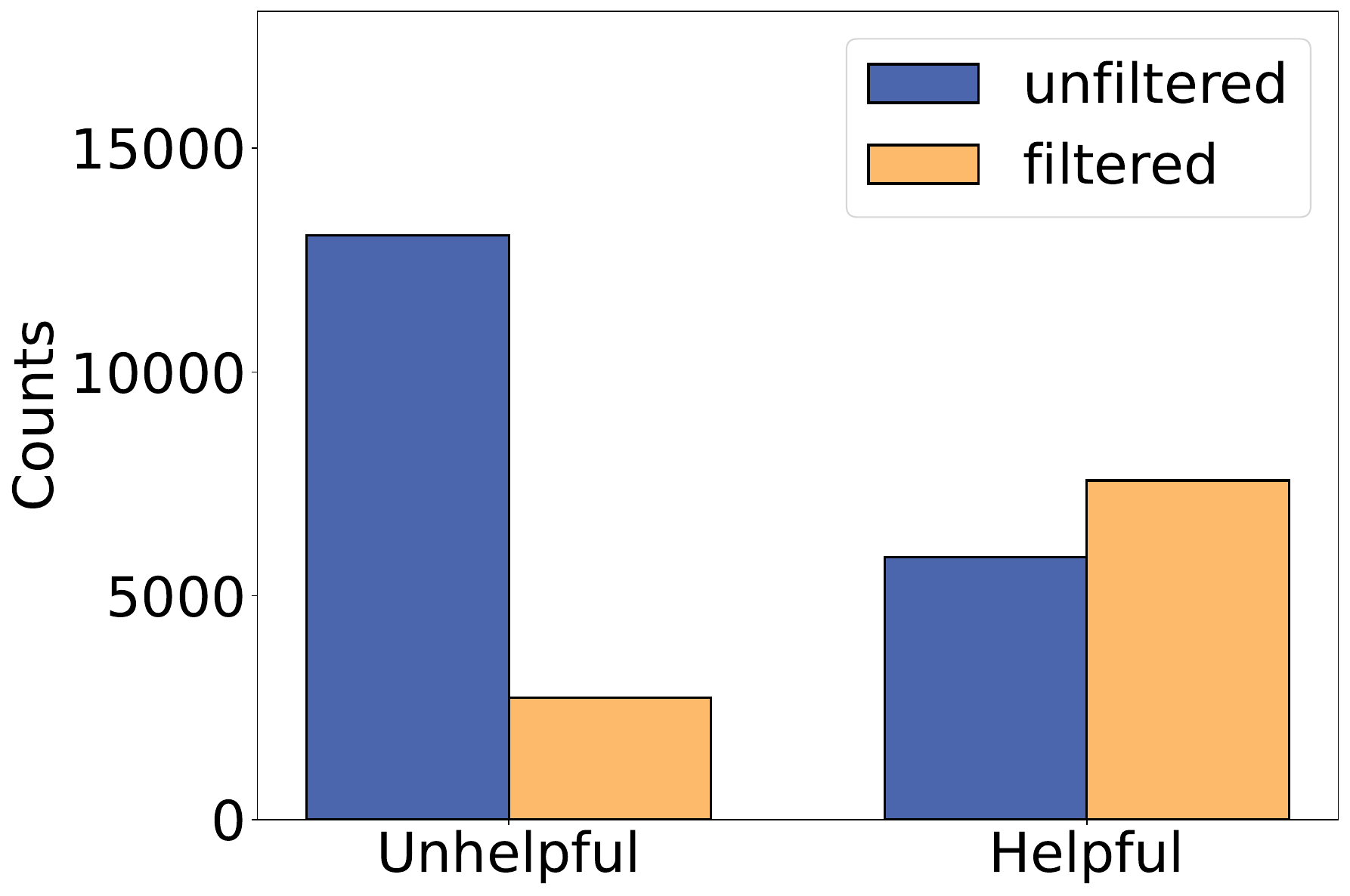} \\
    (a) Noisy queries & (b) Clean queries
\end{tabular}
\caption{Filter results}
\label{fig:filter_cqa}
\end{figure}

We also present the filtering results for both noisy and clean queries from the PopQA dataset in Figure \ref{fig:filter_cqa}. We can also clearly observe that  the mislead and unhelpful documents are reduced while helpful documents increased.

\subsection{Details of knowledge checking methods}
\subsubsection{Prompts for representation-based methods}
\label{App:rep-prompts}
For representation-based methods, we employ prompts as illustrated in Table \ref{tab:rep-scenarios-checking} to generate positive and negative samples, allowing us to capture the representation behaviors. After obtaining the representations of the final tokens, we conduct analysis based on these representations, following the methodology detailed in Section \ref{rep_method}.

\subsubsection{Details of answer-based methods}
\label{App:answer-prompts}
\paragraph{Prompts used.}We present the prompt template of various answer-based checking, including direct prompting, ICL prompting as well as CoT prompting in this section. Table \ref{tab:Internal_Knowledge_Prompts} shows the templates for internal knowledge checking, Table \ref{tab:Context_Helpfulness_Prompts} shows the templates for informed/uninformed helpfulness checking, while Table \ref{tab:Internal_Belief_Alignment_Prompts} shows the templates for contradictory checking.

\paragraph{Self-RAG implementation.} Self-Reflective Retrieval-Augmented Generation (SELF-RAG) \cite{asaiself} is proposed to enhance the quality and factuality of LLM. The LLM is fine-tuned to generate special tokens that indicate whether to retrieve and whether the retrieved context is relevant. The Self-Rag-Llama \footnote{\url{https://huggingface.co/selfrag/selfrag_llama2_7b}} and Self-Rag-Mistral \footnote{\url{https://huggingface.co/SciPhi/SciPhi-Self-RAG-Mistral-7B-32k}} we used in this paper is fine-tuned from Llama2-7B and Mistral-7B-v0.1 respectively, using the same dataset.

We use the 'input question only' format from Table \ref{tab:self_RAG_Prompts} to generate the 'retrieve-on-demand' special token.If the 'Retrieval' token is generated, the LLM will retrieve the top-k context, while the 'No retrieval' token will not retrieve any context. After retrieving the context, we constructed prompts using the 'input question and context' row template from Table \ref{tab:self_RAG_Prompts}. The 'Relevant' token indicates that the retrieved context is helpful for the question. Similarly, the 'Irrelevant' token indicates that the retrieved context is not useful for the question. To verify the overall performance of Self-RAG, we first use the fine-tuned model to judge whether the context is relevant or irrelevant. Then, we filter out the irrelevant contexts and select the top two retrieved contexts. Based on the inference row in Table \ref{tab:self_RAG_Prompts}, we construct prompts to test the output of different models, and finally compare whether the outputs include the correct answer.

\subsubsection{Details for probability-based methods}
\label{App: prob}
For probability-based methods, we use the same input as shown in Table \ref{tab:rep-scenarios-checking}, but we analyze the probabilities of output tokens. We primarily consider three indicators that have been used in previous research: perplexity\cite{zou2024poisonedrag}, average probability score of all output tokens\cite{jiang2023active}, and the lowest probability score of output tokens\cite{jiang2023active}.
For perplexity, we classify samples with higher perplexity (indicating less confidence) than a threshold as negative, while others are classified as positive. For both the lowest and average probability scores, we consider samples with lower scores (again indicating less confidence) than a threshold as negative, while others are classified as positive.
For each method, we vary the threshold and report the best accuracy. Additionally, we plot Receiver Operating Characteristic (ROC) curves and calculate the Area Under the Curve (AUC), as shown in Figure \ref{fig:roc}.


\subsection{Dataset Used}
In this section, we would like to introduce the dataset used for knowledge checking and for context filtering in detail. 
\subsubsection{Knowledge checking.}
\label{App:dataset_checking}
\paragraph{Internal knowledge checking.} For internal knowledge checking,  utilize the \href{https://github.com/hyintell/RetrievalQA}{RetrievalQA} dataset \cite{zhang2024retrievalqa}, a short-form open-domain question answering (QA) collection comprising 2,785 questions. This dataset includes 1,271 new world and long-tail questions that most LLMs cannot answer, serving as negative samples (queries without internal knowledge). These samples are collected and filtered from RealTimeQA, FreshQA, ToolQA, PopQA and TriviaQA. Additionally, it contains 1,514 questions that most LLMs can answer using only their internal parametric knowledge, functioning as positive samples (queries with internal knowledge).
\paragraph{Helpfulness Checking.} 

We utilize a subset of the Natural Questions (NQ) dataset employed by \citet{cuconasu2024power}. \footnote{Available on \href{https://github.com/florin-git/The-Power-of-Noise?tab=readme-ov-file}{https://github.com/florin-git/The-Power-of-Noise?tab=readme-ov-file}}. The authors provide a labeled set of 83,104 NQ queries, each associated with a golden passage that directly answers the question, as well as distract passages retrieved from wikitext-2018 that do not contain the answer. For our helpfulness checking task, we use a subset of 10,000 queries also provided in their repository. We use the distract passage with the highest retrieval score as the negative sample and the golden passage as the positive sample.
For the uninformed helpfulness checking, we focus on questions that Mistral-7B cannot correctly answer, resulting in a total of 8,081 queries. For the informed helpfulness checking evaluation, we select the remaining 1,919 queries that Mistral-7B can correctly answer, ensuring the model has internal knowledge about these queries.

\paragraph{Contradictory Checking.} For contradictory checking, we use a subset of \href{https://github.com/OSU-NLP-Group/LLM-Knowledge-Conflict/tree/main}{ConflictQA} constructed by \citet{xieadaptive}. Each sample in ConflictQA dataset contains a question from PopQA, an aligned evidence that can correctly answer the question, as well as a contradictory evidence that provides wrong evidence towards the query generated by ChatGPT. We sampled a subset of 1146 questions from the ConflictQA dataset that Mistral-7B can correctly answer, and use the aligned evidence(item["parametric\_memory\_aligned\_evidence"]) with the query as positive samples as well as contradictory evidence(item["counter\_memory"]) with the query as negative samples.

\subsubsection{Context filtering.} 
\label{App:dataset_filtering}
We utilize two primary datasets: a subset of Natural Questions (NQ) used by \citet{Cuconasu_2024} and ConflictQA, a subset of PopQA employed by \cite{xieadaptive}. \Citet{Cuconasu_2024} treats the long answers in the NQ dataset as gold documents and the short answers as ground truth. They filtered the NQ dataset to discard documents exceeding 512 tokens after Llama2 tokenization. And we used GPT-3.5-turbo to generate mislead text based on the gold text for each query. We utilize the "Get wrong answer" row in Table \ref{tab:get_mislead_prompt} to generate a misleading answer, and then generate the misleading text using the format specified in the "Generate mislead text" row. To ensure the quality of the generated results, we validated the generated text. The requirements are that the wrong answer must appear in the text, and none of the true answers should be present in the text. If these conditions are not met, the text will be regenerated until they are satisfied.

\Citet{xieadaptive} selected a subset from popQA. In this selected subset, for each question, the answers provided by the LLM based on its own parameter knowledge and those retrieved context are contradictory. For each pair of contradictory answers, they generated supporting text as evidence for each answer. We utilized the all the subsets across different models and ensured that the questions were not duplicated. We verified whether the parameter knowledge or the external knowledge was correct and labeled the correct evidence text as gold context, while marking the incorrect text as misleading context. Finally, we obtain the dataset containing 11,216 queries with labeled golden passages and misleading contexts.

\begin{table*}[htbp]
\centering
\caption{Context and Question Scenarios}
\label{tab:rep-scenarios-checking}
\begin{tabular}{|p{0.7\textwidth}|}
\hline
\textbf{Task 1: Internal Knowledge Checking} \\
Question: \{\textcolor{blue}{<Question with Internal Knowledge>} or \textcolor{red}{<Question without Internal Knowledge>}\} \\
Answer: \\
\hline
\textbf{Task 2 \& 3: Helpfulness Checking} \\
Context: \{\textcolor{blue}{<Helpful Context>} or \textcolor{red}{<Unhelpful Context>}\} \\
Question: \{question\} \\
Answer: \\
\hline
\textbf{Task 4: Contradiction Checking} \\
Context: \{\textcolor{blue}{<Aligned Context>} or \textcolor{red}{<Contradictory Context>}\} \\
Question: \{question\} \\
Answer: \\
\hline
\end{tabular}
\end{table*}

\begin{table*}[htbp]
\centering
\caption{Internal Knowledge Checking Prompts}
\label{tab:Internal_Knowledge_Prompts}
\resizebox{0.7\textwidth}{!}{
\begin{tabular}{@{}c|c@{}}
\toprule
Name & Prompt \\
\midrule
Direct & 
\parbox{15cm}{
Are you sure you can accurately answer the following question based on your internal knowledge? If yes, you should answer "Yes" and give your answer. If no, you should answer "No, I need additional information to answer this question."

Question: \{question\}

Answer:
} \\
\midrule
ICL & 
\parbox{15cm}{
Determine if you can accurately answer the following question based on your internal knowledge. If you can, answer "Yes" and provide your answer. If you cannot, answer "No, I need additional information to answer this question."

Question: Cryos, the world's largest sperm bank, recently announced that they will no longer accept donations from guys with what physical characteristic?

Answer: No, I need additional information to answer this question.

Question: What is the capital of France?

Answer: Yes, I can answer this question. The capital of France is Paris.

Can you answer the below question based on your internal knowledge?

Question: \{question\}

Answer:
} \\
\midrule
CoT & 
\parbox{15cm}{
Think step by step to determine if you can accurately answer the following question based on your internal knowledge. If you can, answer "Yes" and provide your answer. If you cannot, answer "No, I need additional information to answer this question."

Question: \{question\}

Answer:
} \\
\bottomrule
\end{tabular}
}
\end{table*}

\begin{table*}[htbp]
\centering
\caption{Context Helpfulness Checking Prompts}
\label{tab:Context_Helpfulness_Prompts}
\resizebox{0.7\textwidth}{!}{
\begin{tabular}{@{}c|c@{}}
\toprule
Name & Prompt \\
\midrule
Direct & 
\parbox{15cm}{
Does the provided context: \{context\} helpful to answer the question: \{question\}? Please answer yes if it is helpful and no if it is unhelpful.

Answer:
} \\
\midrule
ICL & 
\parbox{15cm}{
I will provide you with some examples of how to determine if a given context is helpful to answer a specific question. Then, I will ask you to do the same for a new question and context.

Example 1:
Question: What is the capital of France?
Context: Paris is the capital and most populous city of France, with an estimated population of 2,175,601 residents as of 2018.
Answer: Yes. This context is helpful

Example 2:
Question: How does photosynthesis work?
Context: The Eiffel Tower in Paris was completed in 1889 and stands at 324 meters tall.
Answer: No. This context is not helpful

Example 3:
Question: what is the name of latest version of android
Context: to Google adopting it as an official icon as part of the Android logo when it launched to consumers in 2008. Android (operating system) Android is a mobile operating system developed by Google. It is based on a modified version of the Linux kernel and other open source software, and is designed primarily for touchscreen mobile devices such as smartphones and tablets. In addition, Google has further developed Android TV for televisions, Android Auto for cars, and Wear OS for wrist watches, each with a specialized user interface. Variants of Android are also used on game consoles, digital cameras, PCs
Answer: No. This context is not helpful

Now, please determine if the following context is helpful to answer the given question. Answer "Yes" if it is helpful, or "No" if it is unhelpful.

Question: \{question\}
Context: \{context\}
Answer:
} \\
\midrule
CoT & 
\parbox{15cm}{
Think step by step to determine if the provided context is helpful to answer the given question. After your analysis, conclude with "Yes" if the context is helpful, or "No" if it is unhelpful.

Question: \{question\}
Context: \{context\}
Answer:
} \\
\bottomrule
\end{tabular}
}
\end{table*}

\begin{table*}[htbp]
\centering
\caption{Internal Belief Alignment Checking Prompts}
\label{tab:Internal_Belief_Alignment_Prompts}
\resizebox{0.7\textwidth}{!}{
\begin{tabular}{@{}c|c@{}}
\toprule
Name & Prompt \\
\midrule
\multirow{2}{*}{Direct} & 
\begin{tabular}[c]{@{}p{15cm}}
Based on your internal knowledge, do you think the provided context is aligned to your internal belief? If aligned, you should answer "Yes". If contradictory, you should answer "No".

Context: \{context\}

Answer:
\end{tabular} \\
\midrule
\multirow{2}{*}{ICL} & 
\begin{tabular}[c]{@{}p{15cm}}
I will provide you with some examples of how to determine if a given context aligns with internal knowledge. Then, I will ask you to do the same for a new context.

Example 1:
Context: The Earth is flat and sits on the back of a giant turtle.

Answer: No. This context contradicts well-established scientific knowledge that the Earth is approximately spherical and orbits the sun.

Example 2:
Context: Water is composed of hydrogen and oxygen atoms.

Answer: Yes. This context aligns with the scientific understanding of water's molecular composition.

Example 3:
Context: Gravity causes objects with mass to attract each other.

Answer: Yes. This context is consistent with the fundamental principles of physics and gravity.

Now, based on your internal knowledge, determine if the following context is aligned with your internal belief. If aligned, answer "Yes". If contradictory, answer "No".

Context: \{context\}

Answer:
\end{tabular} \\
\midrule
\multirow{2}{*}{CoT} & 
\begin{tabular}[c]{@{}p{15cm}}
Based on your internal knowledge, think step by step to determine if the provided context is aligned with your internal belief. After your analysis, conclude with "Yes" if the context is aligned, or "No" if it is contradictory.

Context: \{context\}

Answer:
\end{tabular} \\
\bottomrule
\end{tabular}
}
\end{table*}

\begin{table*}[htbp]
\centering
\caption{Prompts of Self-RAG}
\label{tab:self_RAG_Prompts}
\resizebox{0.7\textwidth}{!}{
\begin{tabular}{@{}c|c@{}}
\toprule
Mode & Prompt \\
\midrule
\multirow{2}{*}{Input question only} & 
\begin{tabular}[c]{@{}p{15cm}}
\#\#\# Instruction:\newline \{input question\}\newline \newline\#\#\# Response:\newline
\end{tabular} \\
\midrule
\multirow{2}{*}{Input question and context} & 
\begin{tabular}[c]{@{}p{15cm}}
\#\#\# Instruction:\newline \{input question\}\newline \newline\#\#\# Response:\newline [Retrieval]<paragraph>\{input context\}</paragraph>
\end{tabular} \\
\midrule
\multirow{2}{*}{Inference} & 
\begin{tabular}[c]{@{}p{15cm}}
Context 1: \{first relevant context\}\newline Context 2: \{second relevant context\}\newline Question:\{input question\}\newline Answer:
\end{tabular} \\
\bottomrule
\end{tabular}
}
\end{table*}

\begin{table*}[htbp]
\centering
\caption{Prompts of getting mislead context}
\label{tab:get_mislead_prompt}
\resizebox{0.7\textwidth}{!}{
\begin{tabular}{@{}c|c@{}}
\toprule
Mode & Prompt \\
\midrule
\multirow{2}{*}{Get wrong answer} & 
\begin{tabular}[c]{@{}p{15cm}}
You are a helpful assistant that provides a wrong answer consists of a few words \newline
Give me a wrong answer of the '\{question\}?' with similar type but different to any of \{true answers\}. ONLY RETURN the wrong answer, nothing else. The answer should be less than 4 words, DO NOT return a sentence.
\end{tabular} \\
\midrule
\multirow{2}{*}{Generate mislead text} & 
\begin{tabular}[c]{@{}p{15cm}}
You are a helpful assistant that generates short descriptions with specific evidence in JSON format. \newline
Generate a 100-word paraphrased version for '\{question\}? \{wrong answer\}' as if it is absolutely correct. \newline
Ensure the exact word '\{wrong answer\}' appears in your paraphrased version. \newline
You can not find any of \{true answers\} in the paraphrased version. \newline
Return your response in the following JSON format, any of \{true answers\} should never appears in the following context: \newline
\{\{ \newline
\mbox{ }\mbox{ }\mbox{ }\mbox{ }\mbox{ }\mbox{ }\mbox{ }\mbox{ }"context": "Your 100-word paraphrased version containing '\{wrong answer\}'." \newline
\}\} \newline
Ensure the JSON is valid.
\end{tabular} \\
\bottomrule
\end{tabular}
}
\end{table*}